\documentclass[twoside,11pt]{article}

\usepackage{jair}
\usepackage{theapa}
\usepackage{epsfig}
\usepackage{amssymb}
\usepackage{times}
\usepackage{subfigure}

\newcommand{\comment}[1]{}
\newtheorem{definition}{\em Definition}
\newtheorem{proposition}{Proposition}
\newtheorem{proof}{Proof}

\input epsf 
\jairheading{27}{2006}{465-503}{04/06}{12/06}
\ShortHeadings{Preference-based Search using Example-Critiquing with
Suggestions} {Viappiani, Faltings, \& Pu} \firstpageno{465}

\begin{document}
\title{Preference-based Search using Example-Critiquing with Suggestions}

\author{\name Paolo Viappiani \email paolo.viappiani@epfl.ch \\
       \name Boi Faltings \email boi.faltings@epfl.ch \\
       \addr Artificial Intelligence Laboratory (LIA)\\
       Ecole Polytechnique F\'ed\'erale de Lausanne (EPFL)\\
       Station 14, 1015 Lausanne, Switzerland
       \AND
       \name Pearl Pu \email pearl.pu@epfl.ch \\
       \addr Human Computer Interaction Group (HCI)\\
       Ecole Polytechnique F\'ed\'erale de Lausanne (EPFL)\\
       Station 14, 1015 Lausanne, Switzerland}

\maketitle

\begin{abstract}
We consider interactive tools that help users search for their most
preferred item in a large collection of options. In particular, we
examine example-critiquing, a technique for enabling users to
incrementally construct preference models by critiquing example
options that are presented to them. We present novel techniques for
improving the example-critiquing technology by adding {\em suggestions} to its
displayed options. Such suggestions
are calculated based on an analysis of users' current preference model and
their potential hidden preferences. We evaluate the performance of
 our model-based suggestion techniques with both synthetic and
real users. Results show that such suggestions are highly attractive to users
and can stimulate them to express more preferences to improve the chance
of identifying their most preferred item by up to $78\%$.
\end{abstract}

\section{Introduction}

The internet makes an unprecedented variety of opportunities
available to people. Whether looking for a place to go for vacation,
an apartment to rent, or a PC to buy, the potential customer is
faced with countless possibilities. Most people have difficulty
finding exactly what they are looking for, and the current tools
available for searching for desired items are widely considered
inadequate. Artificial intelligence provides powerful techniques
that can help people address this essential problem. Search engines
can be very effective in locating items if users provide the correct
queries. However, most users do not know how to map their
preferences to a query that will find the item that most closely
matches their requirements.

{\em Recommender
systems}~\shortcite{Resnick94grouplens,Adomavicius05,ref:burke-survey}
address this problem by mapping explicit or implicit user
preferences to items that are likely to fit these preferences. They
range from systems that require very little input from the users to
more user-involved systems. Many collaborative filtering techniques
~\shortcite{ref:collaborative-filtering}, infer user preferences
from their past actions, such as previously purchased or rated
items.  On the other hand, popular comparison
websites\footnote{E.g., www.shopping.com} often require that users
state at least some preferences on desired attribute values before
producing a list of recommended digital cameras, portable computers,
etc.

In this article, we consider tools that provide recommendations based on explicitly
stated preferences, a task that we call preference-based search. In particular, the problem is defined as:
\begin{quote}
Given a collection ${\cal O} = \{o_1,..,o_n\}$ of $n$ options,
preference-based search (PBS) is an interactive process that helps
users identify the most preferred option, called the {\em target}
option $o_t$, based on a set of preferences that they have stated on
the attributes of the target.
\end{quote}
Tools for preference-based search face a tradeoff between two
conflicting design goals:
\begin{itemize}
\item decision accuracy, measured as the percentage of time that the user
finds the target option when using the tool, and
\item user effort, measured as the number of interaction cycles or task time that
the user takes to find the option that she believes to be the target
using the tool.
\end{itemize}

By target option, we refer to the option that a user prefers most
among the available options. To determine the accuracy of a product
search tool, we measure whether the target option a user finds with
the tool corresponds to the option that she finds after reviewing
all available options in an offline setting. This procedure, also
known as the switching task, is used in consumer decision making
literature~\cite{ref:decision-accuracy}. 
Notice that such procedure is only used to measure the accuracy of a
system. We do not suggest that such procedure models human decision
behavior.

In one approach, researchers focus purely on accuracy in order to
help users find the most preferred choice. For example, Keeney and
Raiffa~\citeyear{ref:keeney-maut} suggested a method to obtain a
precise model of the user's preferences.
This method, known as the value function assessment procedure, asks
the user to respond to a long list of questions. Consider the case
of search for an ideal apartment. Suppose the decision outcome
involves trading off some preferred values of the size of an
apartment against the distance between the apartment and the city
center. A typical assessment question is in the form of ``All else
being equal, which is better: 30 sqm at 60 minutes distance or 20
sqm at 5 minutes distance?'' Even though the results obtained in
this way provide a precise model to determine the most preferred
outcome for the user, this process is often cognitively arduous. It
requires the decision maker to have a full knowledge of the value
function in order to articulate answers to the value function
assessment questions. Without training and expertise, even
professionals are known to produce incomplete, erroneous, and
inconsistent answers~\cite{Tversky1974}. Therefore, such techniques
are most useful for well-informed decision makers, but less so for
users who need the help of a recommender system.

Recently, researches have made significant improvement to this
method. \citeA{ref:chajewska} consider a prior probability
distribution of a user's utility function and only ask questions
having the highest value of information on attributes that will give
the highest expected utility. Even though it was developed for
decision problems under uncertainty, this adaptive elicitation
principle can be used for preference elicitation for product search
which is often modeled as decision with multiple objectives (see in
the related work section the approach
of~\citeR{ref:price-messinger}). \citeA{Boutilier2002} and
\citeA{ref:boutilier-ijcai05} further improved this method by taking
into account the value assigned to future preference elicitation
questions in order to further reduce user effort by modeling the
maximum possible regret as a stopping
criterion.

In another extreme, researchers have emphasized providing
recommendations with as little effort as possible from the users.
Collaborative filtering
techniques~\cite{ref:collaborative-filtering}, for example, infer an
implicit model of a user's preferences from items that they have
rated.  An example of such a technique is Amazon's ``people who 
bought this item also bought..." recommendation. However, users may
still have to make a significant effort in assigning ratings in
order to obtain accurate recommendations, especially as a new user
to such systems (known as the new user problem). Other techniques
produce recommendations based on a user's demographic
data~\cite{Rich79,krulwich97}.

\subsection{Mixed Initiative Based Product Search and Recommender Systems}

In between these two extremes, mixed-initiative dialogue systems
have emerged as promising solutions because they can flexibly scale
user's effort in specifying their preferences according to the
benefits they perceive in revealing and refining preferences already
stated. They have been also referred to as utility and
knowledge-based recommender systems according to
Burke~\citeyear{ref:burke-survey}, and utility-based decision
support interface systems (DSIS) according to Spiekermann and
Paraschiv \citeyear{Spiekermann02}. In a mixed-initiative system,
the user takes the initiative to state preferences, typically in
reaction to example options displayed by the tool. Thus, the user
can provide explicit preferences as in decision-theoretic methods,
but is free to flexibly choose what information to provide, as in
recommender systems.

The success of these systems depends not only on the AI techniques
in supporting the search and recommending task, but also on an
effective user-system interaction model that motivates users to
state complete and accurate preferences. It must strike the right
compromise between the recommendation accuracy it offers and the
effort it requires from the users. A key criterion to evaluate these
systems is therefore the accuracy vs. effort framework which favors
systems that offer maximum accuracy while requiring the same or less
user effort. This framework was first proposed by~\citeA{Payne1993}
while studying user behaviors in high-stake decision making settings
and later adapted to online user behaviors in medium-stake decision
making environments by Pu and Chen~\citeyear{ref:pu-ec05} and Zhang
and Pu~\citeyear{Zhang06}.

In current practice, a mixed-initiative product search and recommender system computes its display set
(i.e., the items presented to the user) based on the closeness of
these items to a user's preference model. However, this set of items
is not likely to provide for diversity and hence may compromise on
the decision accuracy. Consider for example a user who is looking
for a portable PC and gives a low price and a long battery life as
initial preferences. The best matching products are all likely to be
standard models with a 14-inch display and a weight around 3
kilograms. The user may thus never get the impression that a good
variety is available in weight and size, and may never express any
preferences on these criteria. Including a lighter
product in the display set may greatly help a user identify her true
choice and hence increase her decision accuracy.

Recently, the need for recommending not only the best matches,
called the {\em candidates}, but also a diverse set of other items,
called {\em suggestions}, has been recognized. One of the first to
recognize the importance of suggestive examples was
ATA~\cite{linden97interactive}, which explicitly generated examples
that showed the extreme values of certain attributes, called {\em
extreme} examples. In case-based recommender systems, the strategy
of generating both similar and diverse cases was used
\cite{McSherry02,Smyth2003ijcai}. \citeA{ref:diversity-aaai05}
investigated algorithms for generating similar and diverse solutions
in constraint programming, which can be used to recommend
configurable
 products. The complexity
of such algorithms was further analyzed.

So far, the suggestive examples only aim at providing a diverse set
of items without analyzing more deeply whether variety actually
helps users make better decisions. One exception is the
compromise-driven diversity generation strategy by
McSherry~\citeyear{ref:McSherry-compromise} who proposes to suggest
items which are representative of all possible compromises the user
might be prepared to consider. As Pu and Li~\citeyear{ref:pu-ec05}
pointed out, tradeoff reasoning (making compromises) can increase
decision accuracy, which indicates that the compromise-driven
diversity might have a high potential to achieve better decision
quality for users. However, no empirical studies have been carried
out to prove this.

\subsection{Contribution of Our Work}

We consider a mixed-initiative framework with an explicit preference
model, consisting of an iterative process of showing examples,
eliciting critiques and refining the preference model. Users are
never forced to answer questions about preferences they do not yet
possess. On the other hand, their preferences are {\em volunteered and constructed}, not directly
asked. This is the key difference between  navigation-by-proposing used in the mixed-initiative user interaction model as opposed to value assessment-by-asking used in traditional decision support systems.

With a set of simulated and real-user involved experiments, we
argue that including diverse suggestions among the examples shown
by a mixed initiative based product recommender is a significant
improvement in  the state-of-the-art in this field. More specifically,
we show that the model-based suggestion techniques that we have developed
indeed motivate users to express more preferences and help
them achieve a much higher level of decision accuracy without
additional effort.

The rest of this article is organized as follows. We first describe
a set of {\em model-based} techniques for generating suggestions in
preference-based search. The novelty of our method includes: 1) it
expands a user's current preference model, 2) it generates a set of
suggestions based on an analysis of the likelihood of the missing
attributes, and 3) it displays {\em suggested options} whose
attractiveness stimulates users' preference expression. To validate
our theory, we then examine how suggestion techniques help users
identify their target choice in both simulation environments and
with real users. We base the evaluation of these experiments on two
main criteria. Firstly, we consider the completeness of a user's
preference model as measured by preference enumeration, i.e., the
number of features for which a user has stated preferences. The
higher the enumeration, the more likely a user has considered all
aspects of a decision goal, and therefore the decision is more
likely to be rational. Secondly, we consider decision accuracy as
measured by the contrary of the switching rate, which is the number
of users who did not find their target option using the tool and
choose another product after reviewing all options in detail. The
smaller the switching rate, the more likely a user is content with
what she has chosen using the tool, and thus the higher decision
accuracy.

The success of the suggestion techniques is confirmed by
experimental evaluations. An online evaluation was performed with
real users exploring a student housing database. A supervised user
study was additionally carried out with 40 users, performed in a
within-subject experiment setup that evaluated the quantitative
benefits of model-based suggestion. The results demonstrate that
model-based suggestion increased decision accuracy by up to $78\%$,
while the user's effort is about the same as using the
example-critiquing search tool without suggestions. Such user
studies which consider the particular criteria of accuracy vs.
effort have never been carried out by other researchers for
validating suggestion strategies or optimal elicitation procedures.

Finally, we  end by reviewing related works followed by a conclusion.

\section{Example-critiquing}

In many cases, users searching for products or information are not
very familiar with the available items and their characteristics.
Thus, their preferences are not well established, but {\em
constructed} while learning about the
possibilities~\cite{Payne1993}. To allow such construction to take
place, a search tool should ask questions with a complete and
realistic context, not in an abstract way.

\begin{figure}

\centerline{\psfig{file=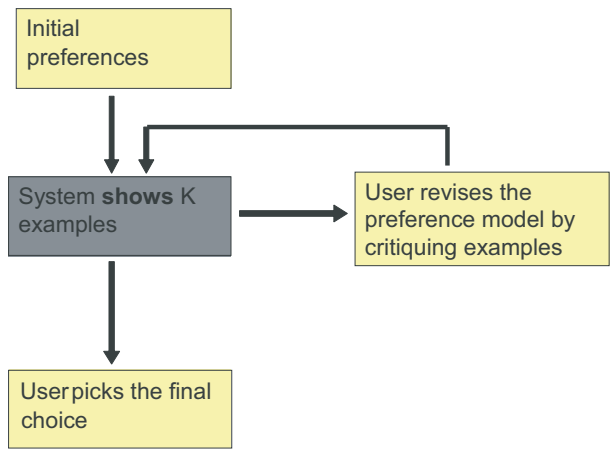,width=8cm}}

\caption{\small Example-critiquing interaction. The dark box is the
computer's action, the other boxes show actions of the user.}
\label{fig:example-critiquing}
\end{figure}

A good way to follow this principle is to implement an {\em example
critiquing} interaction (see Figure~\ref{fig:example-critiquing}).
It shows examples of available options and invites users to state
their critique of these examples. This allows users to better
understand their preferences.

Example-critiquing has been proposed by numerous
researchers in two main forms: systems without and with explicit preference
models:
\begin{itemize}
\item in systems without preference models, the user proceeds by {\em tweaking}
the current best example (``I like this but cheaper'',``I like this
but French cuisine'') to make it fit with his or her preferences
better. The preference model is represented implicitly  by the
currently chosen example and the interaction is that of
navigation-by-proposing. Examples of such systems are the FindMe
systems~\cite{Burke97findme,ref:burke-findme-journal}, the
ExpertClerk system~\cite{ShimazuExpertClerk}, and the dynamic
critiquing systems~\cite{ReillyMMS04}.

\item in systems with preference models, each critique is added to an explicit preference
model that is used to refine the query. Examples of systems with
explicit preference models include the ATA
system~\cite{linden97interactive}, SmartClient~\cite{pu00enriching},
and more recently incremental
critiquing~\cite{ref:incremental-critiquing}.
\end{itemize}

In this article, we focus on example-critiquing with an explicit
preference model for the advantage of effectively resolving users'
preference conflicts. Moreover, this approach not only helps users
make a particular choice, but also obtains an accurate preference
model for future purchases or cross-domain recommendations.

\subsection{Example}

As a simple example consider a student looking for housing. Options
are characterized by the following 4 attributes:
\begin{enumerate}
\item rent in Swiss Francs;
\item type of accommodation: room in a shared apartment,
studio, or apartment
\item distance to the university in minutes;
\item furnished/unfurnished.
\end{enumerate}

Assume that the choice is among the following options:
\begin{center}
\begin{tabular}{rllll}
& rent & type-of-accommodation & distance-to-university & furnished
\\ \hline
$o_1$  & 400 &  room & 17 &  yes \\
$o_2$  & 500 &  room & 32 &  yes \\
$o_3$  & 600 &  apartment & 14 & no \\
$o_4$  & 600 &  studio &  5 & no \\
$o_5$  & 650 &  apartment & 32 &  no \\
$o_6$  & 700 &  studio & 2 & yes \\
$o_7$  & 800 &  apartment & 7 & no \\
\end{tabular}
\end{center}
Assume that the user initially only articulates a preference for the
lowest price. She also has hidden preferences for an unfurnished
accomodation, and a distance of less than 10 minutes to the
university. None of the options can satisfy all of these
preferences, so the most suitable option requires the user to make a
{\em tradeoff} among her preferences. Let us assume that the
tradeoffs are such that option $o_4$ would be the user's most
preferred option. We call this the {\em target} option.

The user may start the search with only the first preference (lowest
price), and the tool would show the $k$ best options according to
the order shown in the table. Here, let $k = 1$ so that only
option $o_1$ is shown.

In an example-critiquing tool without a preference model, the user
indicates a {\em critique} of the currently shown example, and the
system then searches for another example that is as similar as
possible to the current one while also satisfying the critique. In
this case, the user might critique $o_1$ for being furnished, and
the tool might then show $o_3$ which is most similar to the unfurnished
preference. The user might add the critique that the option should be at
most 10 minutes from the university, and the system would then
return $o_7$ as the most similar option that satisfies this
critique. The user might again critique this option as being too
expensive, in which case the system would return to $o_3$ as
most similar to the preference on the "cheaper" option. As there is no memory of earlier
critiques, the process is stuck in a cycle, and the user can never
discover the target $o_4$.

In a tool with a preference model, the user is able to state her
preference for an unfurnished option, making $o_3$ the best option.
Next, she might add the additional preference for a distance of less
than 10 minutes to the university, ending up with $o_4$ which is her
target choice. This illustrates how an explicit preference model
ensures the convergence of the process. In fact, decision theory
shows that when all preferences have been expressed, a user will
always be able to identify the target choice. Note however that more
complex scenarios might require explicit tradeoffs among preferences
to locate the right target choice~\cite{ref:pu-ec04}.

A popular approach to obtain a preference model is to {\em elicit}
it by asking questions to the user. However, this can lead to {\em
means objectives}~\cite{ref:keeney-1992} that distract from the true
target choice. As an example, the tool might first ask the user
whether she prefers a room, a studio or an apartment. If the user
truly has no preference, she might try to translate her preference
for an unfurnished option into a preference for an apartment, since
this is most likely to be unfurnished. However, this is not her true
preference and will shift the best tradeoff from $o_4$ to $o_3$ or
even $o_7$. This illustrates the importance of a mixed-initiative
approach where the user can state preferences in any order on her
own initiative.

The example-critiquing framework raises issues of how to model
preferences, how to generate the solutions shown to the user, and
how to efficiently implement the process. We now briefly summarize
the results of our previous work addressing these issues.

\subsection{Preference Modeling}

When a tool forces users to formulate preferences using particular
attributes or a particular order,  they can fall prey to {\em means
objectives}~\cite{ref:keeney-1992} because they do not have the
catalog knowledge to relate this to their true intentions. Means
objectives are objectives that a person believes to correlate
positively to the true objectives. For example, a manufacturer with
a reputation for good quality may become an objective when it is
impossible to state an objective on the quality itself.

To avoid such means objectives, we require a preference model that
allows users to state preferences incrementally using any attribute,
in any order they wish. Furthermore, the preference model must be
easy to revise at each critiquing cycle by adding or removing
preferences.

This rules out commonly used techniques such as question-answer
dialogues or selection of a fixed set of preferences that are
commonly used on the web today.

An effective formalism that satisfies these criteria is to formulate
preferences using soft constraints. A soft constraint is a function
from an attribute or a combination of attributes to a number that
indicates the degree to which the constraint is violated. More
generally, the values of a soft constraint can be elements of a
semiring~\cite{Bistarelli1997}. When there are several soft
constraints, they are combined into a single preference measure.
Examples of combination operators are summing or taking the maximum.
The overall preference order of outcomes is then given by this
combined measure.

For example, for an attribute that can take values a, b and c, a
soft constraint indicating a preference for value b could map a and
c to 1, and b to 0, thus indicating that only b does not violate the
preference. A preference for the surface area to be at least 30
square meters, where a small violation of up to 5 square meters
could be acceptable, can be expressed by a piecewise linear
function:
\begin{eqnarray*}
1 & & \mbox{if } x < 25 \\
0.2 (30 -x)  & & \mbox{if } 25 \leq x \leq 30 \\
0 & &  \mbox{if } x > 30
\end{eqnarray*}

In example-critiquing, each critique can be expressed as a soft
constraint, and the preference model is incrementally constructed by
simply collecting the critiques. Note that it is also possible for a
user to express several preferences involving the same attributes,
for example to express in one soft constraint that the surface area
should be at least 30 square meters (as above), and in another soft
constraint that it should be no more than 50 square meters. If the
soft constraints are combined by summing their effects, this result
leads to a piecewise linear function:
\begin{eqnarray*}
1 & & \mbox{if } x < 25 \\
0.2 (30 -x)  & & \mbox{if } 25 \leq x \leq 30 \\
0 & &  \mbox{if } 30 < x < 50 \\
0.2 (x -50) & & \mbox{if } 50 \leq x \leq 55 \\
1 & & \mbox{if } x > 55
\end{eqnarray*}
Thus, soft constraints allow users to express relatively complex
preferences in an intuitive manner. This makes soft constraints
 a
useful model for example-critiquing preference models.
 Furthermore,
there exist numerous algorithms that combine branch-and-bound
 with
constraint consistency techniques to efficiently find the most
preferred options in the combined order.
 More details on how to use
soft constraints for preference models are provided by Pu \&
Faltings~\citeyear{ref:constraint-journal-04}.

However soft constraints are a technique that allows a user to
partially and incrementally specify her preferences. The advantage
over utility functions is that it is not necessary to elicit a
user's preference for every attribute. Only attributes whose values
concern the current decision context are elicited. For example, if a
user is not interested in a certain brand of notebooks, then she
does not have to concern herself with stating preferences on those
products. This parsimonious approach is similar to the adaptive
elicitation method proposed by Chajewska et
al.~\citeyear{ref:chajewska}. However, in example-critiquing for
preference-based search, user's preferences are {\em volunteered} as
reactions to the displayed examples, not elicited; users are never
forced to answer questions about preferences without the benefit of
a concrete decision context.

\subsection{Generating Candidate Choices}

In general, users are not able to state each of their preferences
with numerical precision. Instead, a practical tool needs to use an
approximate preference model where users can specify their
preferences in a qualitative way.

A good way to implement such a preference model is to use
standardized soft constraints where numerical parameters are chosen
to fit most users. Such models will necessarily be inaccurate for
certain users. However, this inaccuracy can be compensated by
showing not just one, but a set of $k$ best candidate solutions. The
user then chooses the most preferred one from this set, thus
compensating for the preference model's inaccuracy. This technique
is commonly used in most search engines.

We have analyzed this technique for several types of preference
models: weighted soft constraints, fuzzy-lexicographic soft
constraints, and simple dominance
relations~\cite{Faltings2004solution_generation}.

A remarkable result is that for both weighted and
fuzzy-lexicographic constraint models, assuming a bound on the
possible error (deviation between true value and the one used by the
application) of the soft constraints modeling the preferences, the
probability that the true most preferred solution is within $k$
depends only on the number of the preferences and the error bound of
the soft constraints but not on the overall size of the solution
set. Thus, it is particularly suitable when searching a very large
space of items.

We also found that if the preference model contains many different
soft constraints, the probability of finding the most preferred
option among the $k$ best quickly decreases. Thus, compensating
model inaccuracy by showing many  solutions is only useful when
preference models are relatively simple. Fortunately, this is often
the case in preference-based search, where people usually lack the
patience to input complex models.

As a result, the most desirable process in practice might be a
two-stage process where example-critiquing with a preference model
is used in the first stage to narrow down the set of options from a
large (thousands) space of possibilities to a small (20) most
promising subset. The second phase  would use a tweaking interaction
where no preference model is maintained to find the best choice. Pu
and Chen~\citeyear{ref:pu-ec05} have shown tradeoff strategies in a
tweaking interaction that provide excellent decision accuracy even
when user preferences are very complex.





\subsection{Practical Implementation}

Another challenge for implementing example-critiquing in large scale
practical settings is that it requires solutions to be computed
specifically for the preference model of a particular user. This may
be a challenge for web sites with many users.

However, it has been
shown~\cite{Torrens1998,ref:constraints-journal-2002} that the
computation and data necessary for computing solutions can be coded
in very compact form and run as an applet on the user's computer.
This allows a completely scaleable architecture where the load for
the central servers is no higher than for a conventional web site.
Torrens, Faltings \& Pu~\citeyear{ref:constraints-journal-2002}
describe an implementation of example-critiquing using this
architecture in a tool for planning travel arrangements. It has been
commercialized as part of a tool for business
travelers~\cite{pu00enriching}.

\section{Suggestions}

In the basic example-critiquing cycle, we can expect users to state
any additional preference as long as they perceive it to bring a
better solution. The process ends when users can no longer see
potential improvements by stating additional preferences and have
thus reached an optimum. However, since the process is one of
hill-climbing, this optimum may only be a local optimum. Consider again the example of a user looking for a notebook computer
with a low price range. Since all of the presented products have
about the same weight, say around 3 kg, she might never bother to
look for lighter products. In marketing science literature, this is
called the anchoring effect~\cite{Tversky1974}. Buyers are likely to
make comparisons of products against a reference product, in this
case the set of displayed heavy products. Therefore, a buyer might
not consider the possibility of a lighter notebook that might fit
her requirements better, and accept a sub-optimal result.

Just as in hillclimbing, such local minima can be avoided by
randomizing the search process. Consequently, several authors have
proposed including additional examples selected in order to educate
the user about other opportunities present in the choice of
options~\cite{linden97interactive,ShimazuExpertClerk,McSherry02,ref:smyth-diversity}.
Thus, the displayed examples would include:

\begin{itemize}
\item {\bf candidate} examples that are optimal for the current
preference query, and

\item {\bf suggested} examples that are chosen to stimulate the
expression of preferences.
\end{itemize}

Different strategies for suggestions have been proposed in
literature. Linden~\citeyear{linden97interactive} used extreme
examples, where some attribute takes an extreme value. Others use
diverse examples as
suggestions~\cite{ref:smyth-diversity,Smyth2003ijcai,ShimazuExpertClerk}.

Consider again the example of searching for
housing mentioned in the previous section. Recall that
the choice is among the following options:
\begin{center}
\begin{tabular}{rllll}
& rent & type-of-accommodation & distance-to-university & furnished
\\ \hline
$o_1$  & 400 &  room & 17 &  yes \\
$o_2$  & 500 &  room & 32 &  yes \\
$o_3$  & 600 &  apartment & 14 & no \\
$o_4$  & 600 &  studio &  5 & no \\
$o_5$  & 650 &  apartment & 32 &  no \\
$o_6$  & 700 &  studio & 2 & yes \\
$o_7$  & 800 &  apartment & 7 & no \\
\end{tabular}
\end{center}
In the initial dialogue with the system, the user has stated the
preference of lowest price. Consequently, the options
 are ordered
$o_1 \succ o_2 \succ o_3 = o_4 \succ o_5 \succ o_6 \succ o_7$.

Assume that the system shows only one candidate, which is the
most promising option according to the known preferences: $o_1$.
What other options should be shown as suggestions to motivate
the user to express her remaining preferences?

Linden et al.~\citeyear{linden97interactive} proposed using {\em
extreme} examples, defined as examples where some attribute takes an
extreme value. For example, consider the distance: $o_6$ is the
example with the smallest distance. However, it has a much higher
price, and being furnished does not satisfy the user's other hidden
preference. Thus, it does not give the user the impression that a
closer distance is achievable without compromising her other
preferences. Only when the user wants a distance of less than 5
minutes can option $o_6$ be a good suggestion, otherwise $o_4$ is
likely to be better. Another problem with extreme examples is that
we need two such examples for each attribute, which is usually more
than the user can absorb.

Another strategy~\cite{ref:smyth-diversity,McSherry02,ref:McSherry-compromise,Smyth2003ijcai,ShimazuExpertClerk}
is to select suggestions to achieve a certain diversity, while also
observing a certain goodness according to currently known preferences.
As the tool already shows $o_1$ as the optimal example, the most different
example is $o_5$, which differs in all attributes but does not have an
excessive price. So is $o_5$ a good suggestion? It shows the user
the following opportunities:
\begin{itemize}
\item apartment instead of room: however, $o_3$ would be a cheaper way
to achieve this.
\item distance of 32 instead of 17 minutes: however, $o_2$ would be a
cheaper way to achieve this.
\item unfurnished instead of furnished: however, $o_3$ would be a cheaper
way to achieve this.
\end{itemize}
Thus, while $o_5$ is very diverse, it does not give the user an
accurate picture of what the true opportunities are. The problem is
that diversity does not consider the already known preferences, in
this case price, and the dominance relations they imply on the
available options. While this can be mitigated somewhat by combining
diversity with similarity measures, for example by using a linear
combination of
both~\cite{ref:smyth-diversity,ref:McSherry-compromise}, this does
not solve the problem as the effects of diversity should be limited
to attributes without known preferences while similarity should only
be applied to attributes with known preferences.

We now consider strategies for generating suggestions based on the current preference model. We
call such strategies {\em model-based} suggestion strategies.

We assume that the user is minimizing his or
her own effort and will add preferences to the model only when he or
she expects them to have an impact on the solutions. This is the
case when:
\begin{itemize}
\item the user can see several options that differ in a possible preference, and
\item these options are relevant, i.e. they could be acceptable choices, and
\item they are not already optimal for the already stated preferences.
\end{itemize}
In all other cases, stating an additional preference is irrelevant:
when all options would evaluate the same way, or when the preference
only has an effect on options that would not be eligible anyway or
that are already the best choices, stating it would be wasted
effort. On the contrary, upon display of a suggested outcome
whose optimality becomes clear only if a particular preference is
stated, the user can recognize the importance of stating that
preference. This seems to be confirmed by our user studies.

This has led us to the following principle, which we call the {\em
look-ahead} principle, as a basis for model-based suggestion
strategies:
\begin{quote}
Suggestions should not be optimal under the current preference
model, but should provide a high likelihood of optimality when an
additional preference is added.
\end{quote}
We stress that this is a heuristic principle based on assumptions
about human behavior that we cannot formally prove. However, it is
justified by the fact that suggestion strategies based on the
look-ahead principle work very well in real user studies, as we
report later in this article.

In the example, $o_4$ and $o_3$ have the highest probability of
satisfying the lookahead principle: both are currently dominated by
$o_1$. $o_4$ becomes Pareto-optimal when the user wants a studio,
an unfurnished option, or a distance of less than 14 minutes.
$o_3$ becomes Pareto-optimal when the user wants an apartment, an
unfurnished option, or a distance of less than 17 minutes.
Thus, they give a good illustration of what is possible within
the set of examples.

We now develop our method for computing suggestions and show that how
it can generate these suggestions.

\label{sec:definitions}\subsection{Assumptions about the Preference
Model}

To further show how to implement model-based suggestion strategies,
we have to define preference models and some minimal assumptions
about the shape that user preferences might take. We stress that
these assumptions are only made for generating suggestions. The
preference model used in the search tool could be more diverse or
more specific as required by the application.

We consider a collection of options ${\cal O} = \{o_1,..,o_n\}$ and
a fixed set of $k$ {\em attributes} $A=\{A_1,..,A_k \}$, associated
with domains $D_1,..,D_n$.  Each option $o$ is characterized by the
values $a_1(o),...,a_k(o)$; where $a_i(o)$ represents the value that
$o$ takes for attribute $A_i$.

A {\bf qualitative domain} (the color, the name of neighborhood)
consists in an enumerated set of possibilities; a {\bf numeric
domain} has numerical values (as price, distance to center), either
discrete or continuous. For numeric domains, we consider a function
$range(Att)$ that gives the range on which the attribute domain is
defined. For simplicity we call qualitative (respectively numeric)
attributes those with qualitative (numeric) domains.

 The user's {\em preferences} are assumed to be
independent and
 defined on individual attributes:

\begin{definition}
A {\em preference} $r$ is an order relation $\preceq_{r}$ of the
values of an attribute $a$; $\sim_{r}$ expresses that two values are
equally preferred. A {\em preference model} R is a set of
preferences $\{r_1,..,r_m\}$.
\end{definition}

Note that $\preceq_{r}$ might be a partial or total order.

If there can be preferences over a combination of attributes, such
 as the total travel time in a journey, we assume that the model
 includes additional attributes that model these combinations so that
 we can make the assumption of independent preferences on each
 attribute.
The drawback is that the designer has to know the preferential
dependence in advance. However, this is required for designing the
user interface anyway.

As a preference $r_i$ always applies to the same attribute $a_i$, we
simplify the notation and apply $\preceq_{r_i}$ and $\sim_{r_i}$ to the
options directly: $o_1 \prec_{r_i} o_2$ iff $a_i(o_1) \prec_{r_i}
a_i(o_2)$. We use $\prec_{r_i}$ to indicate that $\preceq_{r_i}$ holds
but not $\sim_{r_i}$.

Depending on the formalism  used for modeling preferences, there are
different ways of combining the order relations given by the
individual preferences $r_i$ in the user's preference model $R$ into
a combined order of the options. For example, each preference may be
expressed by a number, and the combination may be formed by summing
the numbers corresponding to each preference or by taking their
minimum or maximum.

Any rational decision maker will prefer an option to another if the
first is at least as good in all criteria and better for at least
one. This concept is expressed by the {\em Pareto-dominance} (also
just called dominance), that is a partial order relation of the
options.

\begin{definition}
 An option $o$ is {\em Pareto-dominated} by an option $o'$ with respect to $R$ if and only if
 for all  $r_i \in R$, $o \preceq_{r_i} o'$ and for at least one $r_j \in R$,
 $o \prec_{r_j} o'$.  We write $o \prec_{R} o'$ (equivalently we can say that
 $o'$Pareto-dominates $o$ and write $o' \succ_{R} o$).

We also say that $o$ is {\em dominated} (without specifying $o'$).
\end{definition}

Note that we use the same symbol $\prec$ for both individual
preferences and sets of preferences. We will do the same with
$\sim$, meaning that $o \sim_{R} o'$ if $\forall r \in R, o \sim_{r}
o'$.

In the following, the only assumption we make about this combination
is that it is {\em dominance-preserving} according to this
definition of Pareto-dominance. Pareto dominance is the most general
order relation that can be defined based on the individual
preferences. Other forms of domination can be defined as extensions
of Pareto dominance. In the following, whenever we use ``dominance''
without further specification, we refer to Pareto-dominance.

\begin{definition}
A preference combination function is {\em dominance-preserving} if
and only if whenever an option o' dominates another option o in all
individual orders, then o' dominates o in the combined order.
\end{definition}

Most of the combination functions used in practice are
dominance-preserving. An example of a combination that is not
dominance-preserving is the case where the preferences are
represented as soft constraints and combined using {\tt Min()},  as
in fuzzy CSP~\cite{Ruttkay94}. In this case, two options with the
constraint valuations

\begin{itemize}
\item[$o_1$] (0.3, 0.5, 0.7)

\item[$o_2$] (0.3, 0.4, 0.4)
\end{itemize}

 will be considered equally preferred by the combination
function as $Min(0.3, 0.5, 0.7)= 0.3 = Min(0.3, 0.4, 0.4)$, even
though $o_1$ is dominated by $o_2$.

\subsection{Qualitative Notions of Optimality}

The model-based suggestion strategies we are going to introduce are
based on the principle of selecting options that have the highest
chance of becoming optimal. This is determined by considering
possible new preferences and characterizing the likelihood that they
make the option optimal. Since we do not know the weight that a new
preference will take in the user's perception, we must evaluate this
using a qualitative notion of optimality. We present two qualitative
notions, one based only on Pareto-optimality and another based on
the combination function used for generating the candidate
solutions.

We can obtain suggestion strategies that are valid with any
preference modeling formalism, using qualitative optimality criteria
based on the concept of {\em Pareto-dominance} introduced before.

\begin{definition}
 An option $o$ is {\em Pareto-optimal} (PO) if and only if it is not
 dominated by any other option.
\label{def:pareto-opt}
\end{definition}

Since dominance is a partial order, Pareto optimal options can be
seen as the maximal elements of $O$. Pareto-optimality is useful
because it applies to any preference model as long as the
combination function is dominance-preserving.

For any dominance-preserving combination function, an option $o^*$
that is most preferred in the combined preference order is
Pareto-optimal, since any option $o'$ that dominates it would be
more preferred. Therefore, only Pareto-optimal solutions can be
optimal in the combined preference order, no matter what the
combination function is. This makes Pareto-optimality a useful
heuristic for generating suggestions independently of the true
preference combination in the user's mind.

In example-critiquing, users initially state only a subset $R$ of
their eventual preference model $\overline{R}$. When a preference is
added, dominated options with respect to $R$ can become
Pareto-optimal. On the other hand, no option can loose its
Pareto-optimality when preferences are added except that an option
that was equally preferred with respect to all the preferences
considered can become dominated.

Note that one can also consider this as using {\em weak
Pareto-optimality} as defined by Chomicki \citeyear{ref:chomicki},
as we consider that all options are equal with respect to attributes
where no preference has been stated.

We now introduce the notions of \emph{dominating set} and
\emph{equal set}:

\begin{definition}
The dominating set of an option $o$ with respect to a set of
preferences $R$ is the set of all options that dominate $o$:
$O^{>}_{R}(o) = \{ o' \in O : o' \succ_{R} o \}$. We write
$O^{>}(o)$, without specifying $R$, the set of preferences, if $R$
is clear from the context.

The equal set of an option $o$ with respect to $R$ is the set of
options that are equally preferred to $o$: $O^{=}_{R}(o) = \{ o' \in
O : o' \sim_{R} o \}$. We also use $O^{\geq}$ for $O^{>} \cup
O^{=}$.
\end{definition}

The following observation is the basis for evaluating the likelihood
that a dominated option will become Pareto-optimal when a new
preference $r_i$ is stated.

\begin{proposition}\label{prop:pareto-dom}
A dominated option $o$ with respect to $R$ becomes Pareto-optimal
with respect to $R \cup r_i$  if and only if $o$ is
\begin{itemize}
    \item strictly better
with respect to $r_i$ than all options that dominate it with respect
to $R$ and
    \item not worse with respect to $r_i$ than all options that are equally preferred
with respect to $R$.
\end{itemize}
\end{proposition}

\begin{proof}
Suppose there was an option $o'$ that dominates $o$ with respect to
$R$ and that $o$ is not strictly better than $o'$ in the new
preference $r_i$; then $o'$ would still dominate $o$, so $o$ could
not be Pareto-optimal. Similarly, suppose that $o$ is equally
preferred to $o''$ and $o''$ is strictly better than $o$ with
respect to $r_i$; then $o''$ would dominate $o$, so $o$ could not be
Pareto-optimal.
\end{proof}

Thus, the dominating set $O^{>}$ and the equal set $O^{=}$ of a
given option are the potential dominators when a new preference is
considered.

\paragraph{Utility-dominance}

We can consider other forms of dominance as long as they imply
Pareto-dominance. In particular, we might use the total order
established by the combination function defined in the preference
modeling formalism, such as a weighted sum. We call this {\em
utility-domination}, and the utility-optimal option is the most
preferred one.

We may ask when an option can become utility-optimal. A weaker form
of Proposition~\ref{prop:pareto-dom} holds for utility domination:
\begin{proposition}\label{prop:utility-dom}
For dominance-preserving combination functions, a utility-dominated
option $o'$ with respect to $R$ {\em may} become utility-optimal
with respect to $R \cup r_i$ only if $o'$ is strictly better with
respect to $r_i$ than all options that currently utility-dominate it
and not worse than all options that are currently equally preferred.
\end{proposition}
\begin{proof}
Suppose there was an option that became utility-optimal without
being more preferred according to the new preference; then there
would be a violation of the assumption that the combination function
was dominance-preserving.
\end{proof}
Even though it is not a sufficient condition,
Proposition~\ref{prop:utility-dom} can be used as a heuristic to
characterize an option's chance to become utility-optimal.

\subsection{Model-based Suggestion Strategies}

 We propose model-based suggestion strategies that can be
 implemented both with the concept of Pareto- and utility-dominance.
 They are based on the look-ahead principle discussed earlier:
 \begin{quote}
 suggestions should not be optimal under the current preference model,
  but have a high likelihood of becoming optimal when an
additional preference is added.
\end{quote}
We assume that the system knows a subset $R$ of the user's
preference model $\overline{R}$. An ideal suggestion is an option
that is optimal with respect to the full preference model
$\overline{R}$ but is dominated in $R$, the current (partial)
preference model. To be optimal in the full model, from
Propositions~\ref{prop:pareto-dom} and~\ref{prop:utility-dom} we
know that such suggestions have to break the dominance relations
with their dominating set. Model-based strategies order possible
suggestions by the likelihood of breaking those dominance relations.

\subsubsection{Counting Strategy}

The first suggestion strategy, the {\em counting strategy}, is based
on the assumption that dominating options are independently
distributed. From Proposition~\ref{prop:pareto-dom} we can compute
the probability that a dominated option $o$ becomes Pareto-optimal
through a currently hidden preference as:

\begin{displaymath}
p_{opt}(o) = \prod_{o^{'} \in O^{>}(o)} p_d(o,o^{'}) \prod_{o' \in
O^{=}(o)} p_{nw}(o,o')
\end{displaymath}

where $p_d$ is the probability that a new preference makes $o$
escape the domination relation with a dominating option $o'$, i.e.
if $o$ is preferred over $o'$ according to the new
preference;
similarly $p_{nw}$ is the probability that $o$ is not worse than a
equally preferred option $o'$.

 Evaluating this probability requires the exact
probability distribution of the possible preferences, which is in
general difficult to obtain.

The strategy assumes that $p_d = p_{nw}$ is constant for all
dominance relations.
\begin{eqnarray*}
p_{opt}(o) & = & \prod_{o^{'} \in O^{\geq}(o)} p_d\\
 & = & p_d^{|O^{\geq}(o)|}
\end{eqnarray*}

Since $p_d \leq 1$, this probability is largest for the smallest set
$O^{\geq}(o)$. Consequently, the best suggestions are those with the
lowest value of the following counting metric:
\begin{equation}
F_C(o) = |O^{\geq}(o)|
\end{equation}

The counting strategy selects the option with the lowest value of
this metric as the best suggestion.

\subsubsection{Probabilistic Strategy}


The \emph{probabilistic strategy} uses a more precise estimate of
the chance that a particular solution will become Pareto-optimal.

\paragraph{General assumptions}

We assume that each preference $r_i$ is expressed by a cost function
$c_i$. In order to have a well-defined interface, these cost functions
will usually be restricted to a family of functions parameterized by
one or more parameters. Here we assume a single parameter $\theta$,
but the method can be generalized to handle cases of multiple parameters:

\begin{displaymath}
\mbox{ {\bf c}$_i$} = c_i(\theta, a_i(o)) = c_i(\theta,o)
\end{displaymath}

We assume that the possible preferences are characterized by the
following probability distributions:

\begin{itemize}
\item $p_{a_i}$, the probability that the user has a preference
over an attribute $a_i$,

\item $p(\theta)$, the probability distribution of the parameter
associated with the cost function of the considered attribute
\end{itemize}

In the user experiments in the last section, we use a uniform
distribution for both. The probability that a preference on
attribute $i$ makes $o_1$ be preferred to $o_2$ can be computed
integrating over the values of $\theta$ for which the cost of $o_1$
is less than $o_2$. This can be expressed using the Heavyside step function $H(x)\equiv\mbox{\bf if } (x>0) \mbox{ \bf then } 1 \mbox{ \bf else } 0$:

\begin{displaymath}
\delta_i(o_1,o_2)=\int_{\theta} H(c_i(\theta,o_2)-c_i(\theta,o_1))
p(\theta) d\theta
\end{displaymath}
For a qualitative domain, we iterate over $\theta$ and sum up the
probability contribution of the cases in which the value of $\theta$
makes $o_1$ preferred over $o_2$:

\begin{displaymath}
\delta_i(o_1,o_2)=\sum_{\theta \in D_i}
H(c_i(\theta,o_2)-c_i(\theta,o_1)) p(\theta)
\end{displaymath}

To determine the probability of simultaneously breaking the
dominance relation with all dominating or equal options in $O^{\geq}$, a first
possibility is to assume independence between the options, and thus
calculate $\delta_{i}(o,O^{\geq}) = \prod_{o' \in O^{\geq}}
\delta_i(o,o')$, where $\delta_i$  is the chance of breaking one
single domination when the preference is on attribute $i$.

A better estimate can be defined that does not require the
independence assumption, and directly considers the distribution of
all the dominating options. For breaking the dominance relation with
all the options in the dominating set through $a_i$, all dominating
options must have a less preferred value for $a_i$ than that of the
considered option.

For numeric domains, we have to integrate over all possible values of
$\theta$, check whether the given option $o$ has lower cost than all
its dominators in $O^{>}$ and weigh the probability of that
particular value of $\theta$.

\begin{displaymath}
\delta_i(o,O^{>})=
 \int [\prod_{o' \in O^{>}}H(c_i(\theta,o')-c_i(\theta,o)) ]
 p(\theta)d\theta
\end{displaymath}

For qualitative domains, we replace the integral with a
summation over $\theta$.

We also need to consider the second condition of
Proposition~\ref{prop:pareto-dom}, namely that no new dominance relations
with options in the equal set should be created. This can be done by adding
a second term into the integral:

\begin{equation}
\delta_i(o,O^{\geq})=
 \int[\prod_{o' \in O^{>}}H(c_i(\theta,o')-c_i(\theta,o)) \prod_{o'' \in O^{=}}H^{*}(c_i(\theta,o'')-c_i(\theta,o))]
 p(\theta)d\theta \label{eq:general-breakall-dom}
\end{equation}

where $H^{*}$ is a modified Heavyside function that assigns value 1 whenever
the difference of the two costs is 0 or greater. ($H^{*}(x)\equiv\mbox{\bf if }
(x\geq0) \mbox{ \bf then } 1 \mbox{ \bf else } 0$).

We consider the overall probability of becoming Pareto optimal when
a preference is added as the combination of the event that the new
preference is on a particular attribute, and the chance that a
preference on this attribute will make the option be preferred over
all values of the dominating options:
\begin{equation}
F_P(o) = 1 - \prod_{a_i \in A_u} (1 - P_{a_{i}}
\delta_i(o,O^{\geq}))
\end{equation}
If we assume that the user has only one hidden preference, we can
use the following simplification:
\begin{equation}
F_P(o) = \sum_{a_i \in A_u} P_{a_{i}} \delta_{i}(o,O^{\geq})
\end{equation}
which is also a good approximation when the probabilities for
additional preferences are small.
In both cases, we select the options with the highest values as
suggestions.

The computation depends on the particular choice of preference
representation and in many cases it can be greatly simplified by
exploiting properties of the cost functions. In general, the
designer of the application has to consider what preferences the
user can express through the user interface and how to translate
them into quantitative cost functions. A similar approach is taken
by Kiessling \citeyear{KiesslingK02} in the design of {\tt
PREFERENCE SQL}, a database system for processing queries with
preferences.

We now consider several examples of common preference functions and show
how the the suggestions can be computed for these cases.

\paragraph{Preference for a single value in a qualitative domain} 


 Let $\theta$ be the value preferred by the user; the function
$c_i(\theta,x)$ gives a penalty to every value for attribute $a_i$
except $\theta$ . This would allow to express statements like ``I
prefer German cars'', meaning that cars manufactured in Germany are
preferred to cars manufactured in another country.

\begin{displaymath}
c_i(\theta, x) \equiv \mbox{ \bf if } a_i(x)=\theta \mbox{ \bf then
} 0 \mbox{ \bf else } 1.
\end{displaymath}

The probability of breaking a dominance relation between option
$o_1$ and $o_2$ simplifies to the probability that the value of
option $o_1$ for attribute $i$ is the preferred value, when it
differs from the value of $o_2$.

\begin{displaymath}
\delta_i(o_1,o_2)= \left \{
        \begin{array}{ll}
            p[\theta=a_i(o_1)] & \mbox{ {\bf if} } a_i(o_1) \neq a_i(o_2)\\
            0 & \mbox{{\bf otherwise}} \\
        \end{array}
        \right .
        \label{eq-calcdelta-qual-semplified}
\end{displaymath}

Assuming a uniform distribution, $p(\theta)= \left .
        \begin{array}{c}
                1\\
            \hline
                |D_i| \\
        \end{array}
        \right .$ for any $\theta$ (meaning that any value
        of the domain is equally likely to be the preferred value), the probability becomes $1/|D_i|$ when $a_i(o_1) \neq a_i(o_2)$, and 0 otherwise.


The probability of breaking all dominance relations with a set of dominators
without creating new dominance relations is the same as that for a single
dominator, as long as all these options have a different value for $a_i$:
\begin{eqnarray}
 \delta_{i}(o,O^{\geq}) \left \{
        \begin{array}{ll}
            1/|D_i| & \mbox{{\bf if}}\, (\forall o' \in O^{>})\; a_i(o) \neq a_i(o') \\
            0 & \mbox{{\bf otherwise}} \\
        \end{array}
    \right .
\end{eqnarray}

Note that, given the structure of the preference,
$\delta_{i}(o,O^{\geq})=\delta_{i}(o,O^{>})$, because an option $o$
can only break the dominance relations if $a_i(o)$ takes the preferred
value and in that case, no other option can be strictly better
with respect to that preference.

\paragraph{Directional preferences}

\begin{figure}

\centerline{\psfig{file=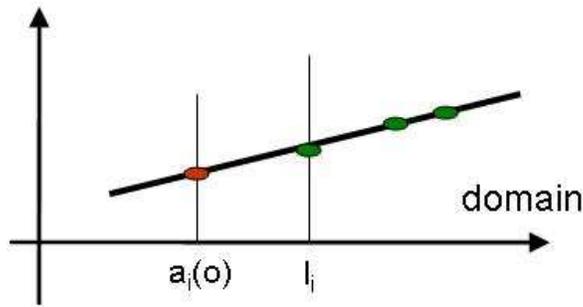,width=8cm}}

\caption{\small In a directional preference, the cost function is
a monotone function of the attribute value. In the case shown here,
smaller values are preferred.} \label{fig:naturalpreferences}
\end{figure}

A particular case of preferences in numeric domains is when the
preference order can be assumed to have a known direction, such as for price (cheaper is always preferred, everything else being equal).
In this case, $\delta(o_1,o_2)$ can be computed by
simply comparing the values that the options take on that attribute
(Figure~\ref{fig:naturalpreferences}).

\begin{eqnarray}
\delta_i(o_1,o_2) \left \{
        \begin{array}{ll}
\mbox{{\bf if} } a_i(o_1) < a_i(o_2) \mbox{ {\bf then} }1\mbox{ {\bf else} }0 & \mbox{  }a_i \mbox{ numeric, natural preference }< \\
\mbox{{\bf if} } a_i(o_1) > a_i(o_2) \mbox{ {\bf then} }1\mbox{ {\bf else} }0 &\mbox{  }a_i \mbox{ numeric, natural preference }>\\
        \end{array}
    \right .
\end{eqnarray}

For a set of options $O^{\geq}$ whose values on
$a_i$ lie between $l_i$ and $h_i$ we have

\begin{eqnarray}
\delta_i(o,O^{\geq}) \left \{
        \begin{array}{ll}
1 \,\mbox{ {\bf if} } a_i(o) < l_i \\
0 \,\mbox{ {\bf otherwise}}  \\
        \end{array}
    \right .
\end{eqnarray}

when smaller values are always preferred, and

\begin{eqnarray}
\delta_i(o,O^{\geq}) \left \{
        \begin{array}{ll}
1 \,\mbox{ {\bf if} } a_i(o) > h_i \\
0 \,\mbox{ {\bf otherwise}}  \\
        \end{array}
    \right .
\end{eqnarray}

when larger values are always preferred. Note that in both cases,
the expressions are independent of the shape of the cost function as
long as it is monotonic.


\paragraph{Threshold preferences in numeric domains}

\begin{figure}

\centerline{\psfig{file=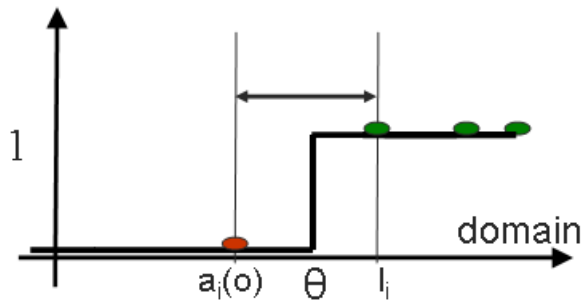,width=8cm}}

\caption{\small \small When the preference {\tt LessThan($\theta$)}
is represented by a step function, an option is preferred over a
set of options with minimum value $l_i$ if the reference value $\theta$
falls in between the values of the given option and $l_i$.}
\label{fig:steppreferences}
\end{figure}

\begin{figure}

\centerline{\psfig{file=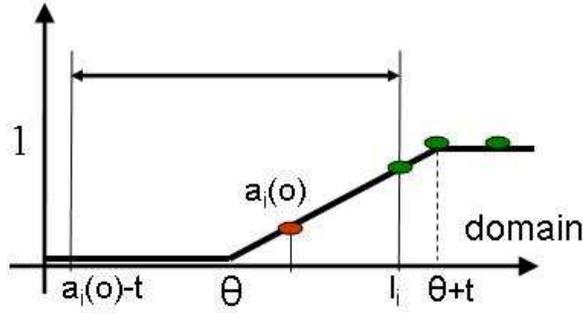,width=8cm}}

\caption{\small When the preference {\tt LessThan($\theta$)}
is represented by a graded step function, an option is preferred over a
set of options with minimum value $l_i$ if the reference value $\theta$
falls in the interval between $a_i(o)-t$ and
$l_i$, where $t = 1/\alpha$.} \label{fig:monotonic-preferences}
\end{figure}

Another commonly used preference expression in numeric domains is
to define a smallest or largest acceptable threshold, i.e. to express
a preference {\tt LessThan($\theta$)} (the value
should be lower than $\theta$) or {\tt GreaterThan($\theta$)} (the
value should be greater than $\theta$). Such a preference is most
straightforwardly expressed by a cost function that follows a step curve~(Figure~\ref{fig:steppreferences}).
To express the fact that there is usually some tolerance for small violations,
more generally a graded step function, where the cost gradually increases,
might be used~(Figure~\ref{fig:monotonic-preferences}).

A possible cost function for {\tt LessThan} might be the following:
\begin{equation}
c_{less-than}(\theta, x) = \left \{
        \begin{array}{ll}
            \mbox{Min}(1, \alpha*(x-\theta)) & \mbox{ {\bf if} } x > \theta\\
            0 & \mbox{{\bf otherwise}} \\
        \end{array}
        \right .
        \label{eq-sample-continous-penalty}
\end{equation}

assigning a penalty when the option takes a value greater than the
reference value $\theta$; such cost is the difference between the
value and the reference, up to a maximum of 1. $\alpha$ is a
parameter that expresses the degree to which the violations can be
allowed; for the following computations it is convenient to use
the length of the ramp from 0 to 1 $t = 1/\alpha$.

In this case the computation of $\delta(o_1,o_2)$ will be, if
$a_i(o_1)<a_i(o_2)$:

\begin{displaymath}
\delta_i(o_1,o_2)=\int_{a_i(o_1)- t}^{a_i(o_2)} 1
p(\theta)d\theta = p[(a_i(o_1)-t) < \theta < a_i(o_2)];
\end{displaymath}

and 0 otherwise (since lower values are preferred in Equation
~\ref{eq-sample-continous-penalty}).

When the transition phase from 0 to 1 is small (the cost function
approximates a step function as in
Figure~\ref{fig:steppreferences}), $\delta_i(o_1,o_2)\simeq
p[a_i(o_1) -t < \theta < a_i(o_2)]$, approximating the probability of
the reference point falling between the two options. Assuming
uniform distribution, the probability evaluates to $(a_i(o_2) -
a_i(o_1) + t)/range(a_i)$, where $range(a_i)$ is that difference
between the largest and smallest values of $a_i$.
The reasoning is illustrated by Figure~\ref{fig:monotonic-preferences}.

The probability computed is conditioned on the knowledge of the polarity
of the user's preference ({\tt LessThan} in this case), and
needs to be weighted by the probability of that polarity. Below,
we assume that both polarities are equally likely, and use a weight of 1/2.



All the dominance relations can be broken simultaneously only if the
considered option has a value for that attribute that is smaller or
bigger than that of all the options in the dominating set. To estimate the
probability that the reference value for the new preference falls in
such a way that all the dominance relations are broken, it is
sufficient to consider the extrema of the values that the dominating
options take on the considered attribute:
\begin{itemize}
\item $h_i=max_{o' \in O^{>}}a_i(o')$
\item $l_i=min_{o' \in O^{>}}a_i(o')$
\end{itemize}

If the values for the current option lies outside the interval
$[l_i, h_i]$, we can consider the probability of breaking all the
relations as in the single dominance case. It will be proportional
to the difference between the current option value and the
minimum/maximum, scaled by the range of values for $a_i$:

\begin{eqnarray}
\delta_{i}(o,O^{\geq}) \left \{
        \begin{array}{ll}
             (a_i(o_1) - h_i+t)/2*range(a_i) &
             \mbox{if $a_i(o_1) > h_i$ }\\
            &\\
            (l_i - a_i(o_1)+t)/2*range(a_i) &
              \mbox{if $a_i(o_1) < l_i$ } \\
            &\\
            0 & \mbox{otherwise} \\
            \end{array}
    \right .
\end{eqnarray}

\paragraph{Peaked preferences for numeric domains}

\begin{figure}

\centerline{\psfig{file=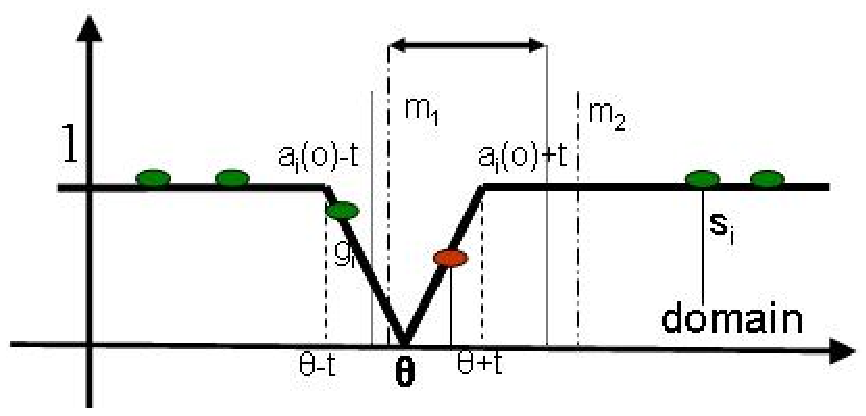,width=8cm}}

\caption{\small An example of peaked preferences. $g_i$ is the
greatest value below $a_i(o)$ of $a_i$ for any option in $O^{\geq}(o)$,
$s_i$ is the smallest value above $a_i(o)$.
$m_1=(a_i(o)+g_i)/2$, $m_2=(a_i(o)+s_i)/2$ are the two midpoints
between $a_i(o)$ and $g_i$, $s_i$. To make $o$ be preferred over all
options in $O^{\geq}(o)$, $\theta$ has to fall between $max(m_1,a_i(o)-t)$
and $min(m_2,a_i(o)+t)$. As it can be seen graphically, in this case
the interval is $]m_1, a_i(o)+t[$.} \label{fig:peakedpreferences}
\end{figure}

Another common case is to have preferences for a particular
numerical value $\theta$, for example {\em ``I prefer to arrive around 12am''}.
To allow some tolerance for deviation, a cost function might have a
slope in both directions:

\begin{displaymath}
c_{peak}(x,\theta) = \alpha*|a_i(o)-\theta|.
\end{displaymath}

In this case, an option is preferred to another one if it is closer
to $\theta$. For example, letting $m$ be the midpoint between
$a_i(o_1)$ and $a_i(o_2)$ and supposing $a_i(o_1) < a_i(o_2)$, we
have

\begin{displaymath}
\delta(o_1,o_2) = p[\theta < m]
\end{displaymath}

For calculating the probability of simultaneously breaking all the
dominance relations without generating new ones, we define $g_i$ as the maximum of all dominating or equal options with a value for $a_i$ less than $a_i(o)$ and $s_i$ as the minimum value of all dominating or equal options greater than $a_i(o)$. As option $o$ is more preferred whenever $a_i(o)$ is closer to $\theta$,
and the interval for $\theta$ where this is the case is one half the interval
between $s_i$ and $g_i$, we have:
\[
\delta(o,O^{\geq})=\frac{s_i-g_i}{range(a_i)}
\]

A more realistic cost function would include a ``saturation point'' from which the cost always evaluates to 1, as shown in Figure~\ref{fig:peakedpreferences}:
\begin{equation}
c_{peak-with-saturation}(x,\theta) = \mbox{Min}(1,
\alpha*|a_i(o)-\theta|).
\end{equation}
Let $t=1/\alpha$ be the tolerance of the preference to either side,
$g_i$ be the greatest value below $a_i(o)$ of $a_i$ for any option in $O^{\geq}(o)$, and $s_i$ be the smallest value above $a_i(o)$.
We define two midpoints $m_1=(a_i(o)+g_i)/2$ and $m_2=(a_i(o)+s_i)/2$,
and we then have:
\begin{displaymath}
\delta(o,O^{\geq})=p[\mbox{max}(m_1, a_i(o)-t)<\theta<\mbox{min}(m_2,
a_i(o)+t)]
\end{displaymath}
If the reference point is uniformly distributed, this evaluates to:
\begin{equation}
\delta(o,O^{\geq})= \frac{\min(m_2,a_i(o)+t) - \max(m_1, a_i(o)-t)}{range(a_i)}
\end{equation}

\subsection{Example}

The following table shows the relevant values for the example
shown earlier. Recall that we had earlier identified $o_4$ and
$o_3$ as the most attractive suggestions.

\begin{center}
\begin{scriptsize}
\begin{tabular}{rlllllllll}
& $O^+$ & rent & type & $\delta_2$ & distance &
$\delta_3$ & furnished & $\delta_4$  & ${\bf p_{opt}}$ \\
& & ($a_1$) & ($a_2$) & & ($a_3$) & & ($a_4$) & & \\
\\ \hline
$o_1$ & - & 400 &   room & - & 17 & - &  yes & - & - \\
$o_2$ & $o_1$ & 500 &   room & 0 & 32 & 0.25 &  yes & 0 & 0.125 \\
$o_3$ & $o_1,o_2$ & 600 &  apartment & 0.5 & 14 & 0.05 & no & 0.5 & 0.451 \\
$o_4$ & $o_1,o_2$ & 600 &   studio & 0.5 &  5 & 0.20 & no & 0.5 & 0.494 \\
$o_5$ & $o_1-o_4$ & 650 &   apartment & 0 & 32 & 0 & no & 0 & 0 \\
$o_6$ & $o_1-o_5$ & 700 &   studio & 0 & 2 & 0.05 & yes & 0 & 0.025\\
$o_7$ & $o_1-o_6$ & 800 &  apartment & 0 & 7 & 0 & no & 0 & 0 \\
\end{tabular}
\end{scriptsize}
\end{center}
In the counting strategy, options are ranked according to the size of the
set $O^+$. Thus, we have $o_2$ as the highest ranked suggestion, followed
by $o_3$ and $o_4$.

In the probabilistic strategy, attribute values of an option are
compared with the range of values present in its dominators. For
each attribute, this leads to the $\delta$ values as indicated in
the table. If we assume that the user is equally likely to have a
preference on each attribute, with a probability of $P_{a_i} = 0.5$,
the probabilistic strategy scores the options as shown in the last
column of the table. Clearly, $o_4$ is the best suggestion, followed
by $o_3$. $o_2$ and also $o_6$ follow further behind.

Thus, at least in this example, the model-based strategies are successful
at identifying good suggestions.

\subsection{Optimizing a Set of Several Suggestions}

The strategies discussed so far only concern generating single
suggestions. However, in practice it is often possible to show a set
of $l$ suggestions simultaneously. Suggestions are interdependent,
and it is likely that we can obtain better results by choosing
suggestions in a diverse way. This need for diversity has also been
observed by others~\cite{ShimazuExpertClerk,ref:smyth-diversity}.

More precisely, we should choose a group $G$ of suggested options by
maximizing the probability $p_{opt}(G)$ that at least one of the suggestions
in the set $G$ will become optimal through a new user preference:
\begin{equation}
p_{opt}(G)  = 1 - \prod_{a_i \in A_u} (1 - P_{a_{i}} (1 - \prod_{o'
\in G} (1 - \delta_i(o',O^{\geq}(o')))))
\end{equation}

Explicitly optimizing this measure would lead to combinatorial
complexity. Thus, we use an algorithm that adds suggestions one by
one in the order of their contribution to this measure given the
already chosen suggestions. This is similar to the algorithm used by
Smyth and McClave~\citeyear{ref:smyth-diversity} and by Hebrard et
al.~\citeyear{ref:diversity-aaai05} to generate diverse solutions.

The algorithm first chooses the best single suggestion as the first
element of the set $G$. It then evaluates each option $o$ as to how
much it would change the combined measure $p_{opt}(G)$ if it were
added to the current $G$, and adds the option with the largest increment.
This process repeats until the desired size of set $G$ is reached.

\subsection{Complexity}

Let $n$ be the number of options, $k$ the number of attributes and
$m$ the number of preferences, $d$ the number of dominators, $A_u$
the attributes on which the user did not state any preference.

All three model-based strategies are based on the dominating set of
an option. We use a straightforward algorithm that computes this as
the intersection of the set of options that are better with respect
to individual preferences. There are $m$ such sets, each with at
most $n$ elements, so the complexity of this algorithm is  $O(n^2
m)$.  In general, the dominating set of each option is of size
$O(n)$ so that the output of this procedure is of size $O(n^2)$, so
it is unlikely that we can find a much better algorithm.

Once the dominating sets are known, the counting strategy has
complexity $O(nd)$, while the attribute and probabilistic strategies
have complexity $O(ndk_{u})$, where $k_{u}=| A_u |$  and $k_{u}<k$.
In general, $d$ depends on the data-set. In the worst case it can be
proportional to $n$, so the resulting complexity is $O(n^2)$.

When utility is used as a domination criterion, the dominating set
is composed by the options that are higher in the ranking. Therefore
the process of  computing the dominating set is highly simplified
and can be performed while computing the candidates. However the
algorithm still has overall worst case complexity $O(n^2)$: the the
last option in the ranking has $n-1$ dominators, and so $d=O(n)$.

When several examples are selected according to their diversity, the
complexity increases since the metrics must be recomputed after
selecting each suggestion.

In comparison, consider the  \emph{extreme} strategy, proposed
initially by Linden et al. in ATA~\citeyear{linden97interactive}. It
selects options that have either the smallest or the largest value
for an attribute on which the user did not initially state any
preference. This strategy needs to scan through all available
options once. Its complexity is $O(n)$, where $n$ is the number of
options (the \emph{size} of the catalog). Thus, it is significantly
more efficient, but does not appear to provide the same benefits as
a model-based strategy, as we shall see in the experiments.

Another strategy considered for comparison, that of generating a
maximally diverse set of
options~\cite{ref:diversity-aaai05,ref:smyth-diversity}, has an
exponential complexity for the number of available options. However,
greedy approximations~\cite{ref:diversity-aaai05} have a complexity
of only $O(n^2)$ , similar to our model-based strategies.

The greedy algorithm we use for optimizing a set of several suggestions
does not add to the complexity; once the distances $\delta_i$ have
been computed for each attribute, the greedy algorithm for computing the set
of suggestions has a complexity proportional to the product of the number of
options, the number of attributes, and the square of the number of
suggestions to be computed.
We suspect that an exact optimization would be NP-hard in the number
of suggestions, but we do not have a proof of this.

\section{Experimental Results: Simulations}

\begin{figure}
 \centering
\includegraphics[width=9.5cm]{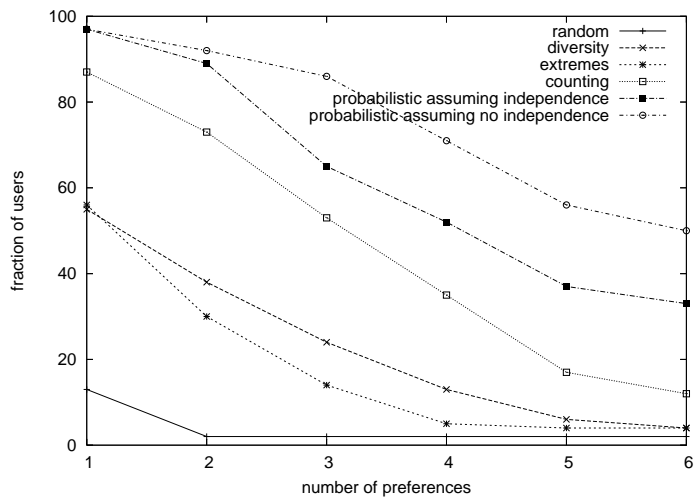}
\caption{\small Simulation results on a database of actual apartment
offers.   For 100 simulated users, each with a randomly chosen
preference model of 6  hidden preferences, we plot the number of
times that the simulation discovered  at least the number of
preferences shown on the abscissa. The higher the curve, the more
preferences were discovered on average.}
\label{fig:simulation-results-unildb}
\end{figure}

\begin{figure}
 \centering
\includegraphics[width=9.5cm]{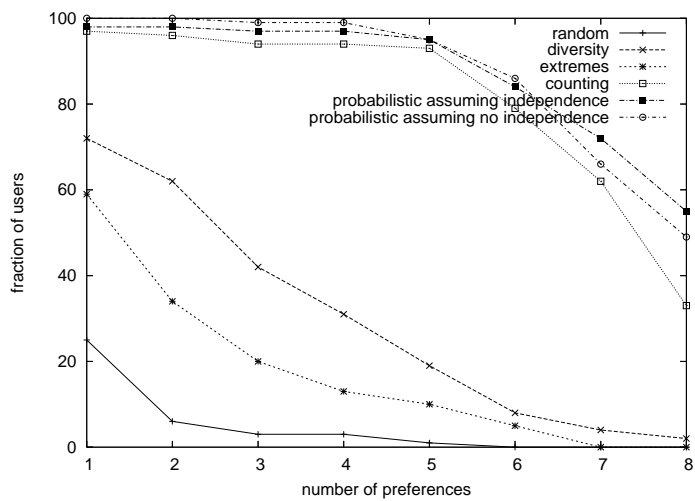}
\caption{\small Simulation results for randomly generated catalogs.
  For 100 simulated users, each with a randomly chosen preference
model of 8 hidden preferences, we plot the number of times that the
simulation discovered  at least the number of preferences shown on
the abscissa. The higher the curve, the more preferences were
discovered on average.} \label{fig:simulation-results-random}
\end{figure}

The suggestion strategies we presented are heuristic, and it is not
clear which of them performs best under the assumptions underlying
their design. Since evaluations with real users can only be carried
out for a specific design, we first select the best suggestion
strategy by simulating the interaction of a computer generated user
with randomly generated preferences. This allows us to compare the
different techniques in much greater detail than would be possible
in an actual user study, and thus select the most promising
techniques for further development. This is followed by real user
studies that are discussed in the next section.

In the simulations, users have a randomly generated set of $m$
preferences on the different attributes of items stored in a
database. As a measure of accuracy, we are interested in whether the
interaction allows the system to obtain a complete model of the
user's preferences. This tests the design objective of the
suggestion strategies (to motivate the user to express as many
preferences as possible) given that the assumptions about user
behavior hold. We verify that these assumptions are reasonable in
the study with real users reported in the next section.

The simulation starts by assigning the user a set of randomly
generated preferences and selecting one of them as an initial
preference. At each stage of the interaction, the simulated user is
presented with 5 suggestions.

We implemented 6 different strategies for suggestions, including the
three model-based strategies described above as well as the
following three strategies for comparison:
\begin{itemize}
\item the {\em random} strategy suggests randomly chosen options;
\item the {\em extremes} strategy suggests options where attributes
take extreme values, as proposed by
Linden~\citeyear{linden97interactive};
\item the {\em diversity} strategy computes the 20 best solutions
according to the current model and then generates a maximally
diverse set of 5 of them, following the proposal of
McSherry~\citeyear{McSherry02}.
\end{itemize}

The simulated user behaves according to an opportunistic model by
stating one of its hidden preferences whenever the suggestions
contain an option that would become optimal if that preference was
added to the model with the proper weight. The interaction continues
until either the preference model is complete, or the simulated user
states no further preference. Note that when the complete preference
model is discovered, the user finds the target option.

We first ran a simulation on a catalog of student accommodations
with 160 options described using 10 attributes. The simulated user
was shown 5 suggestions, and had a randomly generated model of 7
preferences, of which one is given by the user initially. The
results are shown in Figure~\ref{fig:simulation-results-unildb}. For
each value of x, it shows the percentage of runs (out of 100) that
discover at least x out of the 6 hidden preferences in the complete
model. Using random suggestions as the baseline, we see that the
extremes strategy performs only slightly better, while diversity
provides a significant improvement. The model-based strategies give
the best results, with the counting strategy being about equally
good as diversity, and the probabilistic strategies providing
markedly better results.

In another test, we ran the same simulation for a catalog of 50
randomly generated options with 9 attributes, and a random
preference model of 9 preferences, of which one is known initially.
The results are shown in Figure~\ref{fig:simulation-results-random}.
We can see that there is now a much more pronounced difference
between model-based and non model-based strategies. We attribute
this to the fact that attributes are less correlated, and thus the
extreme and diversity filters tend to produce solutions that are too
scattered in the space of possibilities. Also the probabilistic
strategy with both possible implementations (assuming the attributes
values independent or not) give very close results.

\begin{table}
\begin{center}
\begin{tabular}[left]{|c|c|c|c|c|c|c|}
\hline  \#P / \#A  & random & extreme & diversity & counting & prob1 & prob2 \\
\hline 6/6 & 0.12& 0.09& 0.23 & 0.57& 0.59& 0.64\\\hline 6/9 & 0.12& 0.12& 0.27& 0.65& 0.63& 0.67\\\hline 6/12 & 0.11& 0.13& 0.24& 0.62& 0.64& 0.63\\\hline\end{tabular}
\caption{\small The fraction of preferences that are correctly
discovered as a function of the number of attributes; keeping
constant the number of preferences (6) to be discovered. All
attributes have integer domains.} \label{tab:nattrib-comparison}\end{center}\end{table}
\begin{table}\begin{center}\begin{tabular}[left]{|c|c|c|c|c|c|c|}\hline  \#P / \#A & random & extreme & diversity & counting & prob1 & prob2 \\
\hline 3/9 & 0.25& 0.36& 0.28&0.70& 0.71& 0.71 \\\hline 6/9 & 0.11& 0.12& 0.11&0.67& 0.68& 0.68\\\hline 9/9 & 0.041& 0.17& 0.05& 0.66& 0.70&0.73\\\hline\end{tabular}\caption{\small The fraction of preferences that are correctly
discovered (on average) as a function of the number of preferences
to be discovered. All attributes have integer domains.}
\label{tab:npref-comparison}
\end{center}
\end{table}

We investigated the impact of the number of preferences, the number
and type of attributes, and the size of the data set on random data
sets. In the following, {\em prob1} refers to the probabilistic
strategy with the independence assumption, {\em prob2} to the
probabilistic strategy without that assumption.

Surprisingly we discovered that varying the number of attributes
only slightly changes the results. Keeping the number of preferences
constant at 6 (one being the initial preference),  we ran
simulations with the number of attributes equal to 6, 9 and 12. The
average fraction of discovered preferences varied for each strategy
and simulation scenario by no more than $5\%$, as shown in
Table~\ref{tab:nattrib-comparison}.
\begin{table}\begin{center}\begin{tabular}[left]{|c|c|c|c|c|c|c|}\hline  domain & random & extreme & diversity & counting & prob1 & prob2 \\
type     & choice &  &    &     &  &\\
\hline mixed & 0.048& 0.30& 0.18& 0.81& 0.87& 0.86\\

\hline integer &0.04& 0.17& 0.05& 0.66& 0.70& 0.72\\
\hline
\end{tabular}
\caption{\small The fraction of preferences that are correctly
discovered as a function of the different kinds of attribute
domains: integer domains against a mix of 5 integer, 2 discrete
domains and 2 domains with a natural order. We ran 100 simulations
with 9 attributes and 9 preferences.} \label{tab:domain-comparison}
\end{center}
\end{table}

\begin{table}
\begin{center}
\begin{tabular}[left]{|c|c|c|c|c|c|c|}
\hline  data & random & extreme & diversity & counting & prob1 & prob2 \\
size     & choice &  &    &     &  &\\
\hline 50   & 0.25 & 0.50 & 0.56 & 0.89 & 0.94 & 0.93\\
\hline 75   & 0.16 & 0.42 & 0.54 & 0.88 & 0.97 & 0.95\\
\hline 100  & 0.11 & 0.29 & 0.57 & 0.90 & 0.96 & 0.97\\
\hline 200  & 0.05 & 0.22 & 0.54 & 0.86 & 0.91 & 0.93\\
\hline
\end{tabular}
\caption{\small The fraction of preferences that are correctly
discovered as a function of the database size. We ran 100
simulations with 9 attributes and 9 preferences (mixed domains).}
\label{tab:size-comparison}
\end{center}
\end{table}

The impact of the variation of the number of preferences to discover
is shown in Table~\ref{tab:npref-comparison}. All of our model-based
strategies perform significatively better than random choice,
suggestions of extrema, and maximization of diversity. This shows
the importance of considering the already known preferences when
selecting suggestions.

 The performances are higher with mixed domains than with all numeric
domains (Table~\ref{tab:domain-comparison}). This is easily
explained by the larger outcome space in the second case.

Interestingly, as the size of the item set grows, the performance of
random and extreme strategies significantly degrades while the
model-based strategies maintain about the same
performance~(Table~\ref{tab:size-comparison}).

In all simulations, it appears that the probabilistic suggestion
strategy is the best of all, sometimes by a significant margin. We
thus chose to evaluate this strategy in a real user study.

\section{Experimental Results: User Study}

The strategies we have developed so far depend on many assumptions
about user behavior and can only be truly tested by evaluating them
on real users. However, because of the many factors that influence
user behavior, only testing very general hypotheses is possible.
Here, we are interested in verifying that:
\begin{enumerate}
\item using model-based suggestions leads to more complete preference
models.
\item using model-based suggestions leads to more accurate decisions.
\item more complete preference models tend to give more accurate decisions,
so that the reasoning underlying the model-based suggestions is
correct.
\end{enumerate}
We measure decision accuracy as the percentage of users that find
their most preferred choice using the tool. The most preferred
choice was determined by having the subjects go through the entire
database of offers in detail after they finished using the tool.
This measure of decision accuracy, also called the switching rate,
is the commonly accepted measure in marketing
science~\cite<e.g.,>{ref:decision-accuracy}.

We performed user studies using FlatFinder, a web application for
finding student housing that uses actual offers from a university
database that is updated daily. This database was ideal because it
contains a high enough number - about 200 - of offers to present a real
search problem, while at the same time being small enough that it is feasible
to go through the entire list and determine the best choice in less than 1 hour.
We recruited student subjects who
had an interest in finding housing and thus were quite motivated to
perform the task accurately.

We studied two settings:
\begin{itemize}
\item in an unsupervised setting, we monitored user behavior on a publicly
accessible example-critiquing search tool for the listing. This
allowed us to obtain data from over a hundred different users;
however, it was not possible to judge decision accuracy since we
were not able to interview the users themselves.
\item in a supervised setting, we had 40 volunteer students use the
tool under supervision. Here, we could determine decision accuracy
by asking the subjects to carefully examine the entire database of
offers to determine their target option at the end of the procedure.
Thus, we could determine the switching rate and measure decision
accuracy.
\end{itemize}
There are 10 attributes: type of accommodation (room in a family
house, room in a shared apartment, studio apartment, apartment),
rent, number of rooms, furnished (or not), type of bathroom (private
or shared), type of kitchen (shared, private), transportation
available (none, bus, subway, commuter train), distance to the
university and distance to the town center.

For numerical attributes, a preference consists of a relational
operator (less than, equal, greater than), a threshold value and an
importance weight between 1-5; for example, "price less than 600
Francs" with importance 4. For qualitative attributes, a preference
specifies that a certain value is preferred with a certain
importance value. Preferences are combined by summing their weights
whenever the preference is satisfied, and options are ordered so
that the highest value is the most preferred.

Users start by stating a set $P_I$ of initial preferences, and then
they obtain options by pressing a \emph{search} button.
Subsequently, they go through a sequence of {\em interaction cycles}
where they refine their preferences by critiquing the displayed
examples. The system maintains their current set of preferences, and
the user can state additional preferences, change the reference
value of existing preferences, or even remove one or more of the
preferences. Finally, the process finishes with a final set of
preferences $P_F$, and the user chooses one of the displayed
examples.

The increment of preferences $\mid P_F-P_I \mid$ is the number of
extra preferences stated and represents the degree to which the
process stimulates preference expression.

The search tool was made available in two versions:

\begin{itemize}
\item {\bf C}, only showing a set of 6 candidate apartments without
suggestions, and

\item {\bf C+S}, showing a set of 3 candidate apartments and 3
suggestions selected according to the probabilistic strategy with a
utility-dominance criterion.
\end{itemize}

We now describe the results of the two experiments.

\subsection{Online User Study}

\begin{table}
\centerline{
\begin{tabular}{|c|c|c|}
\hline & tool without suggestions & tool with suggestions  \\ \hline
number of critiquing cycles & 2.89 & 3.00 \\
initial preferences & 2.39 & 2.23 \\
final preferences & 3.04 & 3.69 \\
increment & 0.64 & 1.46 \\
\hline
\end{tabular}
} \caption{\small Average behavior of users of the on-line
experiment. We collected logs of real users looking for a student
accommodation with our tool, hosted on the laboratory website.}
\label{tab:online-experiment-results}
\end{table}

FlatFinder has been hosted on the laboratory web-server and made
accessible to students looking for accommodation during the winter
of 2004-2005. For each user, it anonymously recorded a log of the
interactions for later analysis. The server presented users with
alternate versions of the system, i.e. with (\emph{C+S}) and without
(\emph{C}) suggestions. We collected logs from 63 active users who
went through several cycles of preference revision.

In analyzing the results of these experiments, whenever we present a
hypothesis comparing users of the same group, we show its
statistical significance using a paired test. For all hypotheses
comparing users of different groups, we use the impaired student
test to indicate statistical significance. In both cases, we
indicate significance by p, the probability of obtaining the
observed data under the condition that the null hypothesis is true.
Values of p $<$ 0.05 are considered significant, p $<$ 0.01 highly
significant and p $<$ 0.001 very highly significant.

We first considered the increment from initial preference
enumeration $\mid P_I \mid$ to final preference enumeration $\mid
P_F \mid$, as shown in Table~\ref{tab:online-experiment-results}.
This increment was on average 1.46 for the tool with suggestions
\emph{C+S} and only 0.64 for the tool \emph{C} ($128\%$ increase),
showing the higher involvement of users when they see suggestions.
This hypothesis was confirmed with p = 0.002.

It is interesting to see that in both groups the users interacted
for a similar number of cycles (average of 2.89 and 3.00; p = 0.42,
the null hypothesis cannot be rejected), and that the number of
initial preferences is also close (average of 2.39 and 2.23, null
hypothesis cannot be rejected with p = 0.37), meaning that the
groups are relatively unbiased.


The result of the test (Table~\ref{tab:online-experiment-results})
shows clearly that users are more likely to state preferences when
suggestions are present, thus verifying Hypothesis 1. However, as
this is an online experiment, we are not able to
measure decision accuracy. In order to obtain these measures, we
also conducted a supervised user study.

\subsection{Supervised User study}

\begin{table}
\begin{center}
\begin{tabular}{|cc|c|}
 \hline
 \multicolumn{2}{|c|}{Characteristics} & Participants \\ \hline  \hline
 Gender & Male & 31 \\
       & Female & 9 \\ \hline \hline
 Age & 10s & 2 \\
    & 20s & 36 \\
    & 30s & 2 \\ \hline \hline
 Education & Undergraduate & 36 \\
   & Phd & 4 \\ \hline \hline
 \multicolumn{2}{|l|}{Familiar with online apartment search} &  \\
   & Yes & 26 \\
   & No & 14 \\ \hline \hline
 \multicolumn{2}{|l|}{Familiar with apartments in the area} &  \\
    & Yes & 27 \\
   & No & 13 \\ \hline \hline
\end{tabular}
\end{center}
\caption{\small Demographic characteristics of participants for the
supervised user study.} \label{table:demographic}
\end{table}

\begin{table}
\begin{center}
\begin{tabular}{|c|c|c|c|}
  \hline
  \multicolumn{2}{|c|}{} &Interaction with &Interaction with \\
  \multicolumn{2}{|c|}{} &first interface &second interface \\
  \hline
  group 1 & Tool version  & \emph{C} & \emph{C+S} \\ \cline{2-4}
  ({\bf C} first) & Decision Accuracy (mean) & 0.45 & 0.80 \\ \cline{2-4}
   & Preference Enumeration (mean) & 5.30 & 6.15 \\
  \cline{2-4}
                & Interaction cycles (mean) & 5.60 & 4.55 \\ \cline{2-4}
                    & Interaction time (min.,mean) & 8:09 & 4.33 \\
                    \hline
  \hline \hline
  group 2 & Tool version  & \emph{C+S} & \emph{C} \\ \cline{2-4}
  ({\bf C+S} first) & Decision Accuracy (mean) & 0.72 & 0.67 \\ \cline{2-4}
   & Preference Enumeration (mean) & 5.44 & 4.50 \\ \cline{2-4}
                    & Interaction cycles (mean) & 4.05 & 6.25 \\ \cline{2-4}
                    & Interaction time (mean) & 7.39 & 3.33 \\ \hline

\end{tabular}
\end{center}
\caption{\small Results for the supervised experiment. Decision
accuracy and preference enumeration (the number of preferences
stated) are higher when suggestions are provided (interface
\emph{C+S}, showing 3 candidates and 3 suggestions) rather than when
suggestions are not provided (interface \emph{C}, 6 candidates). }
\label{table:supervised-results}
\end{table}

The supervised user study used the same tool as the online user
study but users were followed during their interaction.

To measure improvement of accuracy, we instructed all of the users
to identify their most preferred item by searching the database
using interface 1. This choice was recorded and was called $c_1$.
Then the users were instructed to interact with the database using
interface 2 and indicate a new choice ($c_2$) if the latter was an
improvement on $c_1$ in their opinion. To evaluate whether the
second choice was better than the initial one, we instructed the
users to review all apartments (100 apartments in this case) and
tell us whether $c_1$, $c_2$, or a completely different one truly
seemed best.

Thus, the experiment allowed us to measure decision accuracy, since
we obtained the true target choice for each user. If users stood by
their first choice, it indicated that they had found their target
choice without further help from the second interface. If users
stood by their second choice, it indicated that they had found their
target choice with the help of the second interface. If users chose
yet another item, it indicated that they had not found their target
choice even though they performed search with both interfaces.

40 subjects, mostly undergraduate students, with 9 different
nationalities took part in the study. Most of them (27 out of 40)
had searched for an apartment in the area before and had used online
sites (26 out of 40) to look for accommodations. Table
\ref{table:demographic} shows some of their demographic
characteristics. The subjects were motivated by the interest of
finding a better apartment for themselves, which meant that they
treated the study seriously.

To overcome bias due to learning and fatigue, we divided the users
in two groups, who were asked to interact with the versions in two
different orders:

\begin{itemize}
\item group $1$ used tool \emph{C} (step 1) and then \emph{C+S} (step 2)
\item group $2$ used tool \emph{C+S} (step 1) and then \emph{C} (step 2)
\end{itemize}

Both groups then examined the entire list to find the true most
preferred option. For each version of the tool and each group, we
recorded the fraction of subjects where the final choice made using
that interface was equal to the target option as decision accuracy.
For both groups, we refer to accuracy of interface 1 as $acc_1$, and
accuracy of interface 2 as $acc_2$.

We expected that the order of presenting the versions would be
important. Once the users realized their own preferences and found a
satisfactory option, they are likely to be consistent with that.
Therefore, we expected $acc_2 > acc_1$ in both cases. However, we
expected that average accuracy would significantly increase with
suggestions, and so the results would show $acc_2 >> acc_1$ in the
first group and $acc_2$ only slightly higher than $acc_1$ in group
2.

Table~\ref{table:supervised-results} shows the results. In the next
section we want to verify Hypothesis 2 (decision accuracy improves
with suggestions) and 3 (preference enumeration improves accuracy).
Finally we will check whether a mediation phenomenon is present
(meaning that the improvement of accuracy is entirely explained by
the fact that suggestions lead to an increase of preferences).

\paragraph{Decision Accuracy improves with suggestions}

\begin{figure}
\begin{center}\mbox{
  \subfigure[{\small For group 1, accuracy dramatically increased when they
used the version with suggestions (\emph{C+S}).}]{
    \includegraphics[width=0.49\columnwidth]{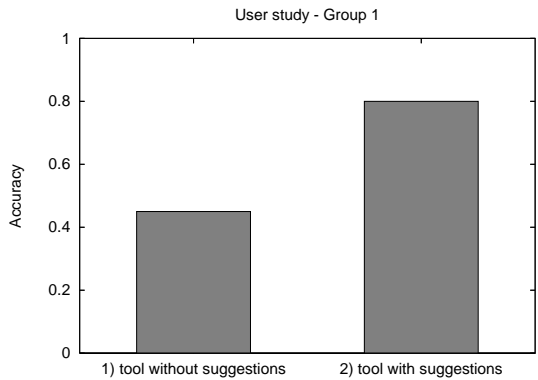}
  }
  \hspace{0.1cm}
  \subfigure[{\small For group 2, accuracy is already very high when they
use the version with suggestions (\emph{C+S}). Further interaction
cycles with the tool \emph{C}  showing 6 candidates does not
increase accuracy any further.}]{
    \includegraphics[width=0.49\columnwidth]{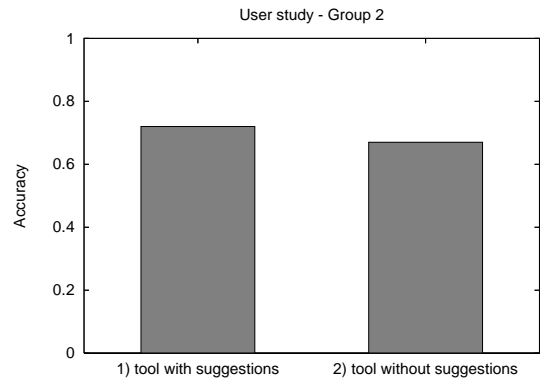}
  }}
\mbox{
  \subfigure[{\small For group 1, users needed less interaction cycles to make a choice
  when using the interface with suggestions (\emph{C+S}).}]{
    \includegraphics[width=0.49\columnwidth]{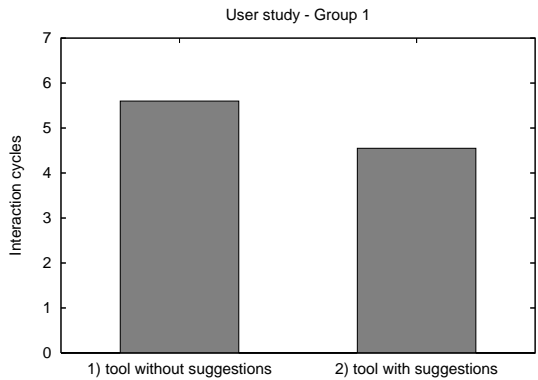}
  }
  \hspace{0.1cm}
  \subfigure[{\small For group 2, the number of interaction cycles significantly increased
  when they used the version without suggestions (\emph{C}).}]{
    \includegraphics[width=0.49\columnwidth]{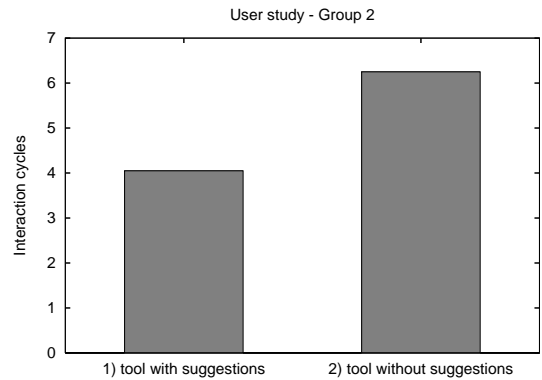}
  }}
  \caption{\small Decision accuracy and interaction cycles for both groups of users of the
  supervised experiment.}
  \label{fig:supervised-experiment-results}
\end{center}
\end{figure}

 Figure~\ref{fig:supervised-experiment-results} shows the variation of decision
 accuracy and the number of interaction cycles for the two groups.

 For group 1, after interaction with tool \emph{C}, the
average accuracy is only 45\%, but after interaction with
\emph{C+S}, the version with suggestions, it goes up to 80\%. This
confirms the hypothesis that suggestions improve accuracy with p =
0.00076. 10 of the 20 subjects in this group switched to another
choice between the two versions, and 8 of them reported that the new
choice was better. Clearly, the use of suggestions significantly
improved decision accuracy for this group.

Users of group 2 used \emph{C+S} straight away and achieved an
average accuracy of 72\% at the outset. We expected that a
consequent use of tool $C$ would have a small positive effect on the
accuracy, but in reality the accuracy decreased to 67\%. 10 subjects
changed their final choice using the tool without suggestions, and 6
of them said that the newly chosen was only equally good as the one
they originally chose. The fact that accuracy does not drop
significantly in this case is not surprising because users remember
their preferences from using the tool with suggestions and will thus
state them more accurately independently of the tool. We can
conclude from this group that improved accuracy is not simply the
result of performing the search a second time, but due to the
provision of suggestions in the tool. Also, the closeness of the
accuracy levels reached by both groups when using suggestions can be
interpreted as confirmation of its significance.

We also note that users of interface \emph{C+S} needed fewer cycles
(and thus less effort) to make decisions (average of 4.15) than
interface \emph{C} (5.92).

 Interestingly, the price of the chosen apartment
increased for the first group (average of 586.75 for \emph{C} to
612.50 for \emph{C+S}; p = 0.04, statistically significant), whereas
it decreased for the second group (average of 527.20 for \emph{C+S}
to 477.25 for \emph{C}; p = 0.18, the decrease is not statically
significant). We believe that subjects in the first group did not
find a good choice, and thus paid a relatively high price to get an
apartment with which they would feel comfortable. Conditioned on
this high price, they were then willing to spend even more as they
discovered more interesting features through suggestions. On the
other hand, subjects in group 2 already found a good choice in the
first use of the tool, and were unwilling to accept a high price
when they did not find a better choice in the second search without
suggestions.

Thus, we conclude that Hypothesis~2 is confirmed: suggestions indeed
increase decision accuracy.

\paragraph{Preference enumeration improves accuracy}

In this study, we notice that when suggestions are present, users
state a higher number of preferences (average of 5.8 preferences vs.
only 4.8 without suggestions, p = 0.021), so that Hypothesis~1 is
again confirmed.

To validate hypothesis 3, that a higher preference enumeration also
leads to more accurate decisions, we can compare the average size of
the preference model for those users who found their target solution
with the first use of the tool and those who did not. In both
groups, users who did find their target in the first try stated on
average 5.56 preferences (5.56 in group 1 and 5.57 in group 2) while
users who did not find their target stated only an average of 4.88
preferences (5.09 in group 1 and 4.67 in group 2). This shows that
increased preference enumeration indeed improves accuracy but
unfortunately we did not find this statistically significant (p =
0.17). In fact, there is a chance that this correlation is due to
some users being more informed and thus making more accurate
decisions and stating more preferences.

\begin{table}
\begin{center}
\begin{tabular}{|c|c|c|}
\hline found &   0.45 &  0.83 \\ \hline
 still not found & 0.55 &   0.17 \\ \hline
 & $\Delta |P|<=0$  & $\Delta |P|>0$ \\
\hline
\end{tabular}
\caption{\small For users who did not find their target in the first
use of the tool, the table shows the fraction that did and did not
find their target in the next try, depending on whether the size of
their preference model did or did not increase. ($\Delta |P|$ is the
variation of the number of stated preferences $|P|$ between the two
uses of the tool).} \label{table:pref-accuracy2}
\end{center}
\end{table}
As an evaluation independent of user's a priori knowledge, we
considered those users who did not find their target in the first
try only. As a measure of correlation of preference enumeration and
accuracy, we considered how often an increase in preference
enumeration in the second try led to finding the most preferred
option on the second try. Table~\ref{table:pref-accuracy2} shows
that among users whose preference model did not grow in size, only
$45\%$ found their target, whereas of those that increased their
preference model, $83\%$ found their target. Again, we see a
significant confirmation that higher preference enumeration leads to
a more accurate decision with real users (p = 0.038251).

\begin{table}
\begin{center}
\begin{tabular}{|c|c|c|c|}
\hline
 $\Delta acc>0$ & 0.23 & 0.14 & 0.38 \\
\hline
 $\Delta acc=0$ & 0.62 & 0.71 & 0.62 \\
\hline
 $\Delta acc<0$ & 0.15 & 0.14 & 0.00 \\
\hline
 {\bf } & $\Delta |P|<0$ & $\Delta |P|=0$ & $\Delta |P|>0$ \\
 \hline
\end{tabular}\caption{\small Variation of accuracy against variation of the
number of stated preferences $|P|$ between the two uses of the
tool.} \label{table:pref-accuracy1}
\end{center}
\end{table}
Finally, a third confirmation can be obtained by considering the
influence that variations in the size of the preference model have
on decision accuracy, shown in Table~\ref{table:pref-accuracy1}.
Each column corresponds to users where the size of the preference
model decreased, stayed the same, or increased. It also shows the
fraction for which the accuracy increased, stayed the same or
decreased (note that when accuracy is 1 at the first step, it cannot
further increase). We can see that a significant increase in
accuracy occurs only when the size of the preference model
increases. In all other cases there are some random variations but
no major increases. The statistical test shows that the hypothesis
that an increase in preference enumeration causes an increase in
accuracy is confirmed with p = 0.0322.

Thus, we conclude that hypothesis 3 is also validated by the user
study; a more complete preference model indeed leads to more
accurate decisions.

\paragraph{Mediation analysis}

Since our three hypotheses are verified, the presence of suggestions
lead to an increase of the preferences stated and consequently to an
increase in accuracy. With a 3-step mediation analysis we want to
check whether there is a mediation phenomenon, meaning that the
increase of accuracy is entirely explained by the increase of the
preferences.

However, a Sobel test did not show statistical significance
(p=0.14), so we cannot conclude that the increase of the preference
enumeration is a ``mediator''. Our interpretation is that
suggestions influence decision accuracy by also making the users
state better preferences.

\subsection{Other Observations}

A more objective measure of confidence is the price that people
are willing to pay for the chosen option as a measure of their
satisfaction, since they would only pay more if the choice satisfies
them more based on the other attributes. For the 40 subjects, the
average rent of the chosen housing with suggestion was CHF 569.85,
an increase of about $7\%$ from the average without suggestions,
which was CHF532.00. In fact, we can observe a general correlation
between price and accuracy, as 9 out of the 10 subjects that did not
find their target in the first interaction finally chose an
apartment with higher rent.

All subjects notably liked the interaction (average 4.1 out of 5)
with no significant difference between the versions.
%
We asked the subjects which version they considered more productive.
The majority of them, 22 out of 40, preferred the version with
suggestions, while 13 preferred the version with more candidates and
5 had no opinion.

Another indication that suggestions are helpful is the average time to
complete the decision task: while it took subjects an average of
8:09 minutes to find their target without suggestions, the version
with suggestions took only 7:39 minutes on average. Thus, using
suggestions users take less time but obtain a more accurate decision.

\section{Related Work}

\paragraph{Example-based search tools}

Burke and others \citeyear{Burke97findme} have been among the first
to recognize the challenge of developing intelligent tools for
preference-based search. Their approach, called {\em assisted
browsing} combines searching and browsing with knowledge based
assistance and recognizes that users are an integral part of the
search process.

They developed the {\em FindMe approach}, consisting of a family of
prototypes that implement the same intuition in a variety of domains
(restaurants, apartments, cars, video, etc.). The main features are the
possibility of similarity based retrieval (look for a restaurant
similar to this, but in San Francisco), the support for tweaking
(look for bigger, nicer, closer to centre, ..), abstraction of high
level features (users might look for a restaurant with {\em casual}
look, where look is not defined in the database directly, but
decoupled into a few basic features), and multiple similarity
metrics. The display follows a hierarchical sort where the
preferences (described by goals: minimize price, find a seafood
cuisine) have a fixed priority. The restaurant advisor was tested
on-line for several years.

Another early and similar work is the {\em ATA} system of Linden et
al.~\citeyear{linden97interactive}. ATA is a tool for planning
travel itineraries based on user's constraints. It followed the
so-called candidate-critiquing cycle where users could post
constraints on their travel and would be shown the 3 best matching
flights from the database. ATA was tested on-line for several
months.

In more recent work,  Shearin and
Lieberman~\citeyear{shearin01intelligent}, have described {\em
AptDecision}, an example-critiquing interface where the user is able
to guide the search by giving feedback on any feature (in the form
of either positive or negative weights) at any time. All of these
critiques are stored in a profile that is displayed at the bottom
part of the interface and can be modified or stored for later use.
Instead of providing feedback manually, the user might prefer to let
AptDecision learn his or her profile weights by comparing two sample
examples. However, they did not investigate strategies for
suggestions.

\paragraph{Improving example selection}

Techniques to induce users to state their preferences more
accurately have been proposed in various recommender systems.
Suggestion mechanisms include extreme values, diversity, and
compound critiques.

The ATA system of Linden et al.~\citeyear{linden97interactive}
included a suggestion strategy of showing extreme examples applied
to the airplane travel domain, for example the first and last flight
of the day. In our simulations, we compared our model-based
techniques to this strategy.

Several researchers
\cite{ref:bridge-diversity,ref:smyth-diversity,McSherry02,McGinty2003iccbr,Smyth2003ijcai,ref:McSherry-compromise}
have studied the issue of achieving a good compromise between
generating similar and diverse results in case-based retrieval. They
consider the problem of finding cases that are most similar to a
given query case,  but at the same time maximize the diversity of
the options proposed to the user. Smyth et.
al~\citeyear{Smyth2003ijcai} improves the common query \emph{show me
more like this}: their adaptive search algorithm alternates between
a strategy that privileges similarity and one that privileges
diversity (\emph{refocus}). McSherry~\citeyear{McSherry02} took this
idea further and provided selection algorithms that maximize
diversity and similarity at the same time.
McSherry~\citeyear{ref:McSherry-compromise} proposes a technique
where retrieved cases are associated with a set of {\em like} cases
that share identical differences with the query case. The like cases
are not displayed among the examples, but accessible to users on
demand. Thus, the retrieval set can be more diverse.

Reilly et al.~\citeyear{ReillyMMS04} also uses a mixture of
similarity and diversity, with the goal of providing possible
standardized critiques to allow trade-offs analysis in an e-commerce
environment. A critique is, in this scope, a modification of a
user's current preferences for narrowing down the search or it is an
indication of a trade-off. Users can select either unit critiques
which revise preferences on individual attributes, or compound
critiques which revise preferences on multiple attributes. The
compound critiques are organized into categories and displayed in
natural language form, for example {\em more memory and larger and
heavier}. One of the innovations in their work is the automatic
generation of sensible critiques involving several features based on
available items using the {\em Apriori} algorithm. Both simulated
and real user studies have shown that compound critiques
significantly reduce the number of interaction cycles.

All of these approaches, however, differ from ours in the sense that
they do not have an explicit preference model. The recent work of
Hebrard et al.~\citeyear{ref:diversity-aaai05} has investigated the
computational problem of generating diverse solutions to constraint
satisfaction problems.

\paragraph{Dialogue-based approaches}

Many other related works try to simulate human conversation in order
to guide the customer through the decision making process.
Shimazu~\citeyear{ShimazuExpertClerk} describes ExpertClerk, an
agent system that imitates a human salesclerk. In the first phase,
the agent tries to narrow down the possibilities by asking
questions. An optimal discrimination tree is built using information
gain (as in ID3 algorithm) where each node represents a specific
question to the user, and the user's answer leads into a specific
portion of the subtree. In fact, each node is equivalent to a crisp
constraint, and the problem of getting to a node with no compatible
examples may occur. In the second phase, the agent proposes three
possible items, chosen to be one in the central and two in the
opposite extreme region of the available product space. It is shown
that an intelligent use of both strategies (asking and proposing) is
more efficient that one of the two strategies alone.

\citeA{Thompson04} also propose a conversational, dialogue-based
approach in {\em ADAPTIVE PLACE ADVISOR}, a conversational
recommendation system for restaurants in the Palo Alto area. Their
approach mimics a conversation that proceeds with questions like
{\em What type of food would you like?}; the user might either
answer with a particular answer like {\em Chinese}, say that he or
she does not care about this aspect, or ask the advisor about the
possible choices. User preferences obtained during the current
conversation are treated as crisp constraints and only items that
satisfy them are considered. When there are no items that satisfy
all preferences, the system may ask the user whether he or she is
willing to relax some constraints.

The tool also develops a long-term user model that keeps track of
preferences expressed in previous interactions. It is used to sort
the results that are shown to the user.

\paragraph{Using prior knowledge}

It is also possible to optimize the set of examples given an
expectation of the user's preferences, without actually asking the
users to state their own preferences. This is the approach described
by Price and Messinger~\citeyear{ref:price-messinger}. This work
differs from ours in that they do not consider preferences of an
individual user, but average preferences for a group of users.

Preference elicitation can be optimized using prior distributions of
possible preferences. This approach was proposed by Chajewska
et~al.~\citeyear{ref:chajewska} to produce a more efficient
preference elicitation procedure. The elicitation is a
question-answering interaction where the questions are selected to
maximize the expected value of information.
Boutilier~\citeyear{Boutilier2002} has extended this work by taking
into account values of future questions to further optimize decision
quality while minimizing user effort. He views the elicitation
procedure itself as a decision process and uses observable Markov
process (POMDP) to obtain an elicitation strategy.

Such approaches require that users are
familiar enough with the available options to answer any question about
value functions without the benefit of example outcomes to assess
them. In contrast, in a mixed-initiative system as described here the
user is free to furnish only the information she is confident about.
It is also questionable whether one can assume a prior distribution
on preferences in personalized recommendation systems where users
may be very diverse.

\section{Conclusion}

We considered AI techniques used for product search and recommender
systems based on a set of preferences explicitly stated by users.
One of the challenges recognized in this field is the elicitation of
an accurate preference model from each user. In particular, we face
the dilemma of accuracy at the cost of user effort.

Some systems may introduce severe errors into the model because
users cannot expend the amount of effort required to state
preferences, while others may require little effort but provide very
general recommendations because the preference model was never
completely established. The ideal solution is one that provides
users with accurate recommendations while minimizing their effort in
stating preferences. Therefore, this article also examined user
interaction issues and emphasized models that motivate users to
state more complete and accurate preferences, while requiring the
least amount of effort from the user.

We conjectured that the benefit of discovering attractive
recommendations presents a strong motivation for users to state
additional preferences. Thus, we developed a model-based approach
that analyzes the user's current preference model and potential
hidden preferences in order to generate a set of suggestions that
would be attractive to a rational user. This suggestion set is
calculated based on the look-ahead principle: a good suggestion is
an outcome that becomes optimal when additional hidden preferences
have been considered. Through simulations, we demonstrated the
superior performance of these model-based strategies in comparison
to the other proposed strategies.

We further validated our hypothesis that such strategies are highly
likely to stimulate users to express more preferences through a
significant within-subject user study involving 40 real users. We
measured decision accuracy, defined as the percentage of users who
actually found their most preferred option with the tool, for an
example-critiquing tool with and without suggestions.


The study showed that users are able to achieve a significantly higher level
of decision accuracy with an example-critiquing tool with suggestions than without suggestions, increasing from
45 to 80$\%$, while the effort spent on both tools is comparable. This shows that there is significant potential
for improving the tools that are currently in use.

It is important to note that this performance is obtained with users
who are not bound to a particular dialogue, but are free to interact
with the system on their own initiative.

This process particularly supports preference expression for users
who are unfamiliar with the domain, and typically for decisions
which require low to medium financial commitments. For highly
important decisions where users understand their preferences well,
other preference elicitation
techniques~\cite{ref:keeney-maut,ref:boutilier-ijcai05} are likely
to provide superior results.

As the strategies are based on the very general notion of Pareto-optimality,
they can be applied to a broad range of preference modeling formalisms,
including utility functions, soft constraints ~\cite{Bistarelli1997}, and
CP-networks ~\cite{Boutilier2004}. This will greatly strengthen the
performance of example-critiquing systems in applications ranging from
decision support to e-commerce.

\section{Acknowledgements}

The authors would like to thank Vincent Schickel-Zuber for his
significant contribution in the development of the web based
interface for FlatFinder, and Jennifer Graetzel for her insightful
suggestions on the various draft versions of this manuscript to
improve its readability. This work was supported by the Swiss
National Science Foundation under contract No. 200020-103421.

\newcommand{\etalchar}[1]{$^{#1}$}
\bibliography{jair}
\bibliographystyle{theapa} 

\end{document}